%% file: main.tex
\PassOptionsToPackage{prologue,dvipsnames}{xcolor}
\documentclass{article}

\usepackage[final]{neurips_2025}

\usepackage{fancyhdr}
\usepackage[utf8]{inputenc} 
\usepackage[T1]{fontenc}    
\usepackage{url}            
\usepackage{booktabs}       
\usepackage{amsfonts}       
\usepackage{nicefrac}       
\usepackage{microtype}      
\usepackage{xspace}
\usepackage[most]{tcolorbox}
\usepackage{subcaption}
\usepackage{soul}
\usepackage{adjustbox}
\usepackage{multirow}
\usepackage{booktabs}       
\usepackage{amsfonts}       
\usepackage{nicefrac}      
\usepackage{microtype}            
\usepackage{colortbl}
\usepackage{enumitem}
\usepackage{amssymb}
\usepackage{wrapfig}
\usepackage{courier}
\usepackage{bbm}
\usepackage{listings}
\usepackage{graphicx}
\usepackage{lipsum} 
\usepackage{booktabs}
\usepackage{wrapfig}
\usepackage[textsize=tiny]{todonotes}
\usepackage{algorithm}
\usepackage{etoc}
\usepackage{pifont}
\usepackage{amsmath}
\usepackage{marvosym}
\usepackage{nccfoots}

\etocdepthtag.toc{mtchapter}
\etocsettagdepth{mtchapter}{subsection}
\etocsettagdepth{mtappendix}{none}
\definecolor{commentgreen}{RGB}{0,128,0}
\definecolor{keywordblue}{RGB}{0,0,180}
\lstdefinestyle{mypythonpseudo}{
    language=Python,
    basicstyle=\ttfamily\footnotesize,
    keywordstyle=\color{keywordblue}\bfseries,
    commentstyle=\color{commentgreen},
    stringstyle=\color{red},
    showstringspaces=false,
    breaklines=true,
    frame=none,
    captionpos=b,
    numbers=none,
    escapeinside={(*@}{@*)},
}

\newcommand{\cmark}{\ding{51}}
\newcommand{\xmark}{\ding{55}}
\newcommand{\ours}[0]{{TON}\xspace}

\newcommand{\our}{TON} %

\definecolor{citecolor}{HTML}{0071bc}
\usepackage[pagebackref=false,breaklinks=true,letterpaper=true,colorlinks,citecolor=citecolor,bookmarks=false]{hyperref}
\definecolor{hidden-red}{RGB}{205, 44, 36}
\definecolor{hidden-blue}{RGB}{194,232,247}
\definecolor{hidden-orange}{RGB}{243,202,120}
\definecolor{hidden-green}{RGB}{34,139,34}
\definecolor{hidden-pink}{RGB}{255,245,247}
\definecolor{hidden-black}{RGB}{20,68,106}
\definecolor{purple}{RGB}{144,153,196}
\definecolor{yellow}{RGB}{255,228,123}
\definecolor{hidden-yellow}{RGB}{255,248,203}
\definecolor{tkcolor}{RGB}{224,223,255}
\definecolor{darkblue}{rgb}{0, 0.40, 0.75}
\newcommand{\eg}{\textit{e.g.,}}
\newcommand{\ie}{\textit{i.e.,}}
\hypersetup{colorlinks=true, citecolor=darkblue, linkcolor=darkblue, urlcolor=darkblue}
\tcbset{
  aibox/.style={
    width=\linewidth,
    top=8pt,
    bottom=4pt,
    colback=blue!6!white,
    colframe=black,
    colbacktitle=black,
    enhanced,
    center,
    attach boxed title to top left={yshift=-0.1in,xshift=0.15in},
    boxed title style={boxrule=0pt,colframe=white,},
  }
}

\newcommand{\authorfootnote}[1]{\Footnotetext{}{\textit{#1}}}

\newtcolorbox{AIbox}[2][]{aibox,title=#2,#1}

\title{
Think or Not? Selective Reasoning via\\Reinforcement Learning for Vision-Language Models
}

\author{%
Jiaqi Wang$^{1\dagger}$\enspace
Kevin Qinghong Lin$^{2\dagger}$\enspace
James Cheng\textsuperscript{$1$}
\enspace Mike Zheng Shou\textsuperscript{$2$\Letter}\vspace{1mm} \\
$^1$The Chinese University of Hong Kong \quad $^2$Show Lab, National University of Singapore
}

\begin{document}

\maketitle

\begin{abstract}
Reinforcement Learning (RL) has proven to be an effective post-training strategy for enhancing reasoning in vision–language models (VLMs).
Group Relative Policy Optimization (GRPO) is a recent prominent method that encourages models to generate complete reasoning traces before answering, leading to increased token usage and computational cost.
Inspired by the human-like thinking process—where people skip reasoning for easy questions but think carefully when needed—we pioneer how to enable VLMs to first decide \textit{when reasoning is necessary}.
To realize this, we propose \ours, a two-stage training strategy:
\textbf{(\textit{i})} a supervised fine-tuning (SFT) stage with a simple yet effective “\textbf{thought dropout}” operation, where reasoning traces are randomly replaced with empty thoughts. This introduces a think-or-not format that serves as a cold start for selective reasoning;
\textbf{(\textit{ii})} a GRPO stage that enables the model to freely explore when to think or not, while maximizing task-aware outcome rewards.
Experimental results show that \ours can \textit{reduce the completion length by up to $\mathbf{90\%}$ compared to vanilla GRPO, without sacrificing performance or even improving it}. 
Further evaluations across LLM (GSM8K), VLM (CLEVR, Super-CLEVR, GeoQA), and Agentic (AITZ) tasks—covering a range of reasoning difficulties under both 3B and 7B models—consistently reveal that the \textit{model progressively learns to bypass unnecessary reasoning steps as training advances}.
These findings shed light on the path toward human-like reasoning patterns in RL approaches.
Our code is available at \url{https://github.com/kokolerk/TON}.
\end{abstract}

\authorfootnote{\hspace{-4mm}$\dagger$\,Equal contribution.\hspace{2mm}\Letter\,Corresponding authors.}


\input{sec/intro}
\input{sec/relatedwork}

\input{sec/method}

\input{sec/exps}

\input{sec/conclusion}

\input{sec/ack}
\bibliographystyle{unsrt}
\bibliography{main}
\newpage
\appendix
\input{sec/appendix}

\end{document}

%% file: sec/intro.tex
\section{Introduction}

\textit{‘‘To think or not to think, that is the question.’’}

Reinforcement learning (RL) has recently emerged as a dominant post-supervised fine-tuning (SFT) strategy in vision-language models (VLMs)~\cite{huang2024vlm,lu2025uir1,luo2025guir1,chen2025sft}. Methods like GRPO~\cite{shao2024deepseekmath} have shown promising results in enhancing reasoning capabilities through KL-divergence losses based on rule-driven rewards. 
However, these approaches often lead to unnecessarily long and redundant reasoning processes due to their reliance on full-length generative trajectories~\cite{su2025underthinking, chen2024unlocking,wu2025understanding}.
To address this inefficiency, some works attempt to shorten reasoning chains with rule-based reward penalties~\cite{chen2024think,luo2025adar1,shen2025dast} during the pre-training phase or introduce external control mechanisms, such as in very recent Qwen3~\cite{qwen2.5}. 
Nonetheless, a more natural and scalable solution is to enable the model to \textit{decide when to think}—mirroring how humans modulate cognitive effort in response to task difficulty.

In this work, we begin by presenting empirical evidence that \textit{thinking is not always necessary}. 
In AITZ~\cite{aitz}, we observe that $51\%$ of questions can be answered correctly even when the entire reasoning trace is omitted, resulting in significant savings in thought tokens. This finding underscores the potential of selective reasoning strategies to improve efficiency without sacrificing accuracy.
Secondly, by exploring a simple prompting strategy — allowing the model to skip reasoning steps for easier queries — we observe that even math-enhanced VLMs struggle to adaptively omit redundant thought generation. Instead, they tend to default to a conservative approach, producing full reasoning traces regardless of task difficulty.
This suggests that the ability to “think or not” is not solely determined by reasoning capacity, but should instead be treated as a distinct skill—one that should be explicitly activated through format-following in supervised fine-tuning (SFT) stage.

Motivated by the above observations, we introduce \textbf{\our} (\ie~\textbf{T}hink-\textbf{o}r-\textbf{N}ot), a two-stage training framework featuring a simple yet effective “\textbf{thought dropout}” approach. 
This method explicitly replace reasoning traces with minimal “\texttt{\textbackslash n\textbackslash n}” delimiter and employs SFT to train the model that reasoning can be skipped—thereby enabling the possibility of bypassing reasoning.
A subsequent GRPO stage further refines this {selective-reasoning} policy via self-exploration, rewarding answers without introducing extra regularization. 
As illustrated in Figure~\ref{fig:teaser}, vanilla GRPO consistently generates reasoning sequences regardless of task difficulty. In contrast, our method, TON, adaptively allocates reasoning based on the complexity of the task. For simple tasks (left), TON can bypass unnecessary reasoning and directly provide the answer, reducing $90\%$ token usage. 
For more hard problems (right), it still engages in detailed, step-by-step reasoning to arrive at the correct solution.
To the best of our knowledge, TON is \textit{\textbf{the first work} to study “when to think” in VLM.}


Built on top of \our, we using the Qwen-2.5-VL series and conduct extensive evaluations across the LLM benchmark (GSM8K~\cite{cobbe2021training}), vision-language tasks—spanning counting (CLEVR~\cite{johnson2017clevr}, SuperCLEVR~\cite{li2023super}) as well as mathematical reasoning (GeoQA~\cite{chen2021geoqa}), and the agent task like mobile agent navigation (AITZ~\cite{aitz})—which collectively cover a spectrum of reasoning levels and diverse task settings.
Overall, we find that \ours \textbf{achieves substantial reductions in completion length without compromising performance}—cutting $87\%$ of tokens on CLEVR and $65\%$ on GeoQA. Notably, on the multi-step navigation task AITZ, \ours reduces the average task-level output length from $3.6K$ to $0.9K$ tokens.
Moreover, we observe that \textbf{omitting reasoning traces can even improve performance}: on GeoQA, \ours outperforms the vanilla GRPO baseline by up to $17\%$ in accuracy, demonstrating a “free-lunch” effect where shorter reasoning outperforms or matches longer trajectories.
Comprehensive ablation studies further reveal that \textbf{the skip-thought ratio increases progressively with reward improvements} during training, suggesting the model learns to selectively bypass unnecessary reasoning steps in an adaptive manner.

%% file: sec/relatedwork.tex
\section{Related Works}
\begin{figure*}[t]
    \centering
    \includegraphics[width=\linewidth]{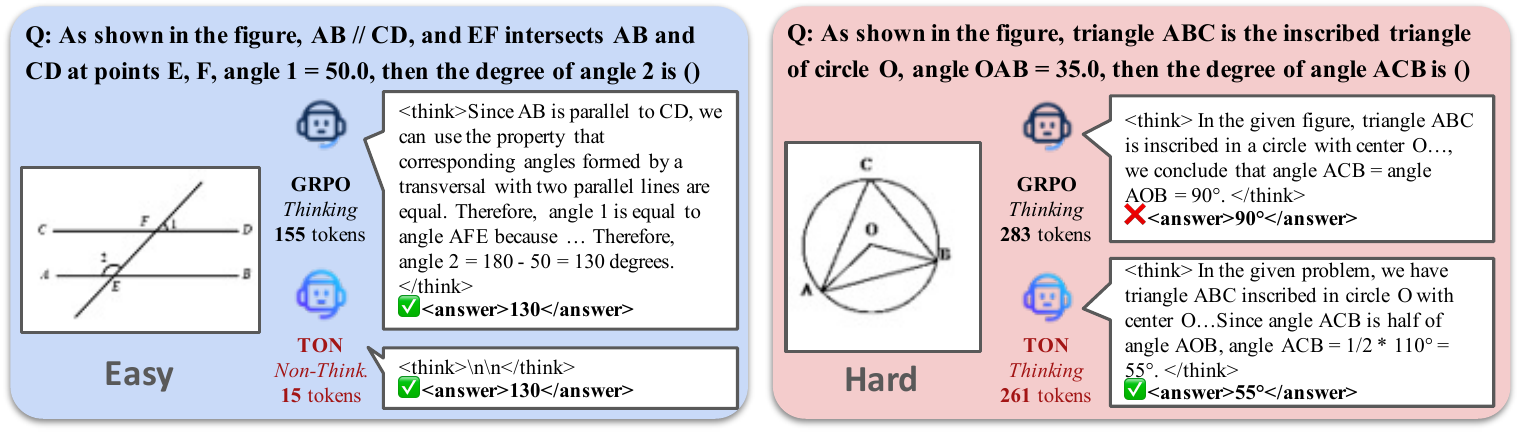}
    \captionsetup{font=small}
    \caption{\textbf{Illustrating the “think or not think” trade-off.}
    \textbf{Left}: For simple queries, explicit reasoning is unnecessary—models like GRPO that always "think" incur redundant computation.
    \textbf{Right}: For more complex geometric problems, step-by-step reasoning is essential to arrive at the correct answer.
    Our proposed TON framework learns to adaptively invoke reasoning only when needed—skipping it for easy cases while engaging in deeper inference for harder tasks.}
    \vspace{-1em}
    \label{fig:teaser}
\end{figure*}

\textbf{Reinforcement Learning for Vision–Language Models.}
Most VLMs start with SFT on large collections of instruction data to acquire broad foundational knowledge~\cite{showui,seeclick,aitz,cogagent}. 
To further improve performance, recent work has adopted a post-training paradigm that leverages human feedback~\cite{gpt4,gemini,deepseekai2025deepseekr1}. 
RL from human feedback (RLHF) fits a reward model on preference annotations and refines the policy via Proximal Policy Optimization (PPO)~\cite{schulman2017proximal,OpenAI2023GPT4V,gemini,vonwerra2022trl}.
Direct Preference Optimization (DPO)~\cite{rafailov2024direct} streamlines this workflow by recasting policy updates as a binary classification task, aligning model outputs distributions with human preferences without a reward module.
Beyond these methods, Group Relative Policy Optimization (GRPO)~\cite{shao2024deepseekmath} blends offline and online learning: it samples groups of thinking process, uses Answer verification (such as Math verifier) as reward feedback, and computes relative advantages within each group. By avoiding a value function, GRPO provide an elegant solution by promoting diverse reasoning paths and improved answer quality. 
Despite a series of GRPO follow-up works~\cite{dapo,drgrpo,chen2024think}, all of these approaches assume that every question demands a full thinking—leading to lengthy decoding. In contrast, our work focuses on {“\textit{when to think}”} instead of {“\textit{how to think}”}: we introduce a selective reasoning policy that learns to skip unnecessary “think” phases, boosting inference efficiency without sacrificing accuracy.

\paragraph{Thinking in Language Models.}
From early Chain-of-Thought~\cite{cotprompt,react,mmreact} prompting to recent reasoning-intensive reinforcement learning approaches~\cite{shao2024deepseekmath,deepseekai2025deepseekr1,bai2024digirl,qi2024webrl}, reasoning has emerged as a core dimension in the development of language models. 
Most existing work emphasizes how to enhance reasoning capabilities, often resulting in increasingly lengthy and complex thought processes~\cite{chen2024unlocking,webagentplan,luo2025adar1} while relatively few studies address the efficiency of reasoning. 
For instance,~\cite{kimi15} proposes a {long2short} strategy to compress decoding length,~\cite{autocurr} encourages models to output “I don't know” to terminate unproductive reasoning, and~\cite{tokenbudget} introduces a token-budget-aware reasoning policy.
While these approaches offer promising insights into controlling reasoning length, we argue for a more foundational perspective: rather than deciding how to reason once the process has started, \textit{models should first determine whether reasoning is necessary at all}. Simple questions may be answered directly without any explicit reasoning, while complex questions may require maintaining a full reasoning trajectory~\cite{wu2025understanding,su2025underthinking,chen2024think}. 
In this work, we explore the selective reasoning paradigm within VLMs by introducing a simple yet effective method -- \textbf{thought-dropout}. We validate its effectiveness on tasks such as Counting, Math, and further extend it to more practical agentic settings.

%% file: sec/method.tex
\section{Preliminary}
\noindent\textbf{Task Definition.}
We formalize the vision-language reasoning environment as a Markov Decision Process (MDP) defined by a tuple $(\mathcal{V}, \mathcal{Q}, \mathcal{S}^*, \pi, r)$, covering a wide range of vision-language tasks. 
Here, $\mathcal{V}$ denotes the visual context (\eg~an image). $\mathcal{Q}$ is a language-based query or question posed about the visual input. 
The model, governed by policy $\pi$, takes the input pair $(\mathcal{V}, \mathcal{Q})$ and generates a predicted answer $\mathcal{{S}}$. 
The environment provides a scalar reward function $r(\mathcal{\cdot})$ 
based on the model’s response $\mathcal{O}$.
A correct prediction, \eg~$\mathcal{O}$ matches the ground truth answer $\mathcal{S}^*$, yields a positive reward, while an incorrect one yields zero.
The objective in this environment is to learn an adaptive policy $\pi_\theta$, parameterized by $\theta$, that maximizes the expected reward, enabling the model to reason selectively and efficiently across diverse input settings.

\noindent\textbf{Reward Function.} 
The reward function $r(\cdot)$ can be either model-based~\cite{schulman2017proximal,vonwerra2022trl} or rule-based, as recently demonstrated in~\cite{shao2024deepseekmath,deepseekai2025deepseekr1}, which is typically categorized into two types: format rewards $r_f$ and outcome rewards $r_o$.
While the outcome rewards are usually carefully designed based on different tasks or requests in previous works~\cite{shao2024deepseekmath,huang2024vlm, lu2025uir1, chen2024think}, the format reward $r_f$, is always shared in the same.
Given the response $\mathcal{O}$, it should follow the 
required HTML tag format \textit{<think>$\mathcal{T}$<\textbackslash think><answer>$\mathcal{S}$<\textbackslash answer>}, where $\mathcal{T}$ is the reasoning process (\ie~a thought) and $\mathcal{S}$ is the predicted answer.
This formulation requires the model to think before deriving the answer and makes it easy to parse both the reasoning process and the final outcome (\eg~via regular expression).

\begin{wrapfigure}{b}{0.4\linewidth}
    \centering
    \vspace{-2em}
    \includegraphics[width=0.6\linewidth]{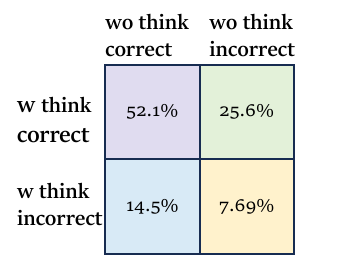} 
    \caption{\textbf{Accuracy comparison of with \textit{v.s.} without “thinking”} during SFT using Qwen-2.5-VL-3B on the AITZ task.
    }
    \vspace{-2em}
    \label{fig: motivation}
\end{wrapfigure}

\section{\ours: Selective Reasoning via Policy Optimization}

\textbf{Observation.}
In practice, humans do not require explicit reasoning for all tasks—many can be completed intuitively. Similarly, models can often produce correct answers to simple questions without explicit thinking.
As illustrated in figure~\ref{fig: motivation}, the percentages of correct and incorrect samples under different setups with and without the thinking process in inference (see Appendix~\ref{app sec: motivation exps} for overall performance). 
We find that $52.1\%$ of answers remained correct without “think,” and $14.5\%$ were even correct only without it—implying that \textit{explicit thinking is not always necessary}.

A straightforward idea is to prompt the model to decide whether to “think” or not (we prompt the model to skip thinking in the simple questions in Sec.~\ref{subsec: prompt vs sft}). However, as shown in our experiments (Figure~\ref{fig: hybrid} and Appendix~\ref{app sec: prompt sft}), the model still tends to generate the full reasoning process without any no-think try.
This suggests that the ability to decide whether to think is not solely governed by reasoning capability, but should instead be treated as a separate skill—one that must be explicitly trained through format-following during the supervised fine-tuning (SFT) stage.
These observations motivate us to activate this ability early in the SFT stage and develop \ours, which enables selective reasoning by automatically switching between “think” and “non-think” modes.

\subsection{First SFT stage: Thought Dropout}
\label{sec: first sft stage}
In the initial stage, the model is typically fine-tuned on “think-answer” formatted data, where the “think” contains high-quality reasoning traces to serve as a cold start.
To extend this predefined reasoning ability to selective reasoning, we view “think” vs. “non-think” as part of the output format itself by \textit{dropping} the “think” component during training.

However, it is difficult to determine which samples should be skipped, as different models exhibit varying reasoning capabilities. Therefore, we begin with \textit{random} dropout and allow the model to learn to decide for itself during the second RL stage (Sec.\ref{rlstage}).
To this end, we propose “\textbf{Thought Dropout}” that randomly injecting empty “thought” segments, requiring only minor code changes:

\begin{algorithm}[h!]
    \caption{Pseudo-code for \texttt{thought\_dropout}}
    \label{alg_thought_dropout}
\begin{lstlisting}[style=mypythonpseudo]
def thought_dropout(thought, dropout_prob):
    if random.random() < dropout_prob:
        thought = "\n\n"
    return thought
\end{lstlisting}
\end{algorithm}

This approach injects both the answer format and the skip-thought format as prior knowledge before the second RL stage.

\noindent\textbf{Where do Thoughts come from?}
Given a policy operating in an environment $(\mathcal{V}, \mathcal{Q}, \mathcal{S}^*, \pi, r)$, a key challenge is how to curate high-quality cold-start “thought” data without relying on external models, such as closed-source APIs.
A naïve approach is to run multiple inference passes and retain only successful cases based on answer matching—but we find this to be less effective.
To address the scarcity of high-quality “thought” data, we instead adopt a \textbf{reverse thinking} strategy: leveraging the base model $\pi$ itself to self-generate a rich corpus of thought sequences.
Specifically, given the visual context $\mathcal{V}$, textual query $\mathcal{Q}$, and ground-truth answer $\mathcal{S}^*$, 
we prompt the policy $\pi_\theta$ to deduce the corresponding intermediate thought as follows: 
\begin{equation}
    \mathcal{T} \;\leftarrow\; \pi_{\theta}\bigl(\mathcal{V}, \mathcal{Q}, \mathcal{S}^*\bigr)
\end{equation}
Specially, we generate intermediate thoughts with the following prompts:
\begin{AIbox}{Prompt for \textit{Reverse Thinking}}
    Based on the following question and image, generate a thought process to explain how to derive the answer from the inputs.\\
     Image: \texttt{\{Image\}} Question: \texttt{\{Question\}} Answer: \texttt{\{Answer\}}\\
    Do not output the answer, only generate the reasoning process. 
    Formulate your outputs using concise language. 

\end{AIbox}

In this way, we curate sufficient thought data without relying on external models. These serve as our cold-start training corpus, enabling us to apply the Thought Dropout strategy during SFT to activate the model’s ability to bypass thoughts.

\subsection{Second RL stage: Group Relative Policy Optimization}
\label{rlstage}
Although SFT teaches the skip-thought format, it still leaves a central question unresolved: when should thoughts be skipped or retained? Ideally, the model should learn to explore this decision on its own.
To this end, we adopt reinforcement learning via GRPO to enhance the model’s ability to explore this decision as part of its reasoning process.

Given an image $v \in \mathcal{V}$ and text query $q \in \mathcal{Q}$, GRPO samples $N$ candidate responses with variations $\{o_1, o_2, \ldots, o_N\}$ from the policy $\pi_\theta$ and evaluates each response $o_i$ using a reward function $r(\cdot)$, 
which measures the quality of the candidate in the context of the given question. 
To determine the relative quality of these responses, GRPO normalizes the rewards by computing their mean and standard deviation and subsequently derives the advantage as:
\begin{equation}
    A_i = \frac{r(o_i) - \text{mean}\{r(o_1) , r(o_2) , \ldots, r(o_N) \}}{\text{std}\{r(o_1) , r(o_2) , \ldots, r(o_N)\}}
\end{equation}
where $A_i$ represents the advantage of the candidate response $o_i$ relative to other sampled responses. GRPO encourages the model to generate responses with higher advantages within the group by updating the policy $\pi_\theta$ using the following objective:
\begin{equation}
\label{eq: grpo}
     \mathcal{J}_{GRPO}(\theta) = \mathbb{E}[{\{o_i\}_{i=1}^N\sim\pi_{\theta_{old}}(v,q)}]\frac{1}{N}\sum_{i=1}^N\left\{\min[\textcolor{black}{\textcolor{Fuchsia}{\alpha_i}\cdot \textcolor{citecolor}{A_i},\ \textcolor{Fuchsia}{\beta_i} \cdot \textcolor{citecolor}{A_i}}]-\beta\mathbb{D}_{KL}[\pi_\theta||\pi_{ref}]\right\}
\end{equation}
\begin{equation}
\label{eq: s1,s2}
    \textcolor{Fuchsia}{{\alpha_{i}}} = \frac{\textcolor{Fuchsia}{\pi_\theta({{o_i}}|v, q)}}{\textcolor{Fuchsia}{\pi_{\theta_{old}}({o_i}|v, q)}}, \quad
    \textcolor{Fuchsia}{\beta_i}  = \text{clip}\left(\frac{\textcolor{Fuchsia}{\pi_\theta({o_i}|v,q)}}{\textcolor{Fuchsia}{\pi_{\theta_{old}}({o_i}|v,q)}},1+\epsilon,1-\epsilon\right). 
\end{equation}

\begin{figure}[t]
    \centering
    \includegraphics[width=\linewidth]{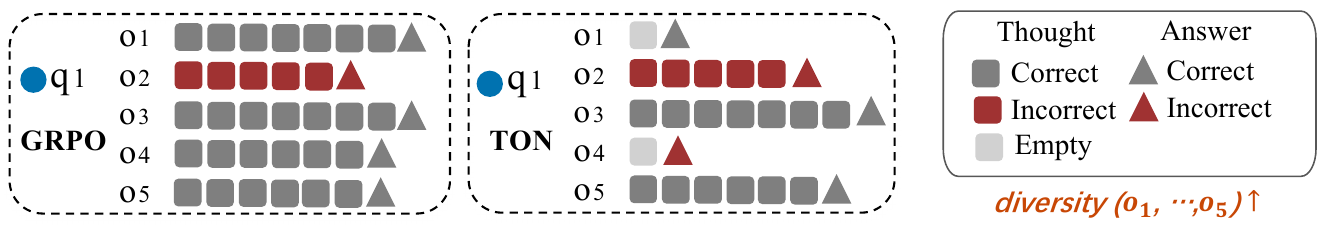}
    \captionsetup{font=small}
    \caption{\textbf{Illustration of the responses from GRPO and \ours.} $q_1$ is the question and $\{o_1, \cdots, o_5\}$ are the generated responses containing thoughts $\mathcal{T}$ (circle) and answers $\mathcal{S}$ (triangle). \ours can sample from the empty think $\mathcal{T}_{\textbackslash n \textbackslash n}$, thus enhancing the response diversity over the vanilla GRPO.}
    \vspace{-5mm}
    \label{fig:ton diagram}
\end{figure}
\textbf{How does TON impact GRPO?} 
As illustrated in Fig.~\ref{fig:ton diagram}, our TON allows the model to choose “empty-think” $\mathcal{T}_{\textbackslash n\textbackslash n}$ during the inference step, thus resulting in a significant variation in the distribution between the non-think ($o_i \sim  \mathcal{T}_{\textbackslash n\textbackslash n}$) and think responses ($o_i \sim\mathcal{T}$) by \ours compared to both think ones ($o_i \sim\mathcal{T}$) generated by vanilla GRPO.
Unlike previous works like DAPO~\cite{dapo} emphasize on advantage distribution $\textcolor{citecolor}{A_i}$ by dynamic sampling in the sparse reward space, our \ours shifts the focus to the latent distribution space of responses ($\textcolor{Fuchsia}{\pi_\theta(o_i|v, q)}$), thus enhancing the diversity of the terms $\textcolor{Fuchsia}{\alpha}$ and $\textcolor{Fuchsia}{\beta}$ in Eq.~\ref{eq: s1,s2}.

\noindent\textbf{How to design Rewards?}
To support GRPO training across diverse settings, it is crucial to carefully examine reward design choices.
We consider two main types of matching:

\textit{(i) Discrete Matching.} 
For tasks with deterministic, categorical or numerical outputs—e.g., classification, counting, or math problems—we use a binary value reward $r_{d}(s, g) = \mathbf{1}(s = g)$: if the predicted answer $s$ matches the ground-truth $g$, we assign $r_{d} = 1$; otherwise, $r_{d} = 0$.

\textit{(ii) Continous Matching}. 
For tasks producing continuous outputs—e.g., spatial coordinates in UI navigation or object grounding—we allow a tolerance region. Given a predicted point \(\mathbf{p}=[x,y]\) and a ground-truth box \(\mathbf{b}=[x_1,y_1,x_2,y_2]\), we define:
\[
  r_{c}(\mathbf{p}, \mathbf{b}) =
  \begin{cases}
    1, & \mathbf{p}\text{ lies inside }\mathbf{b},\\
    0, & \text{otherwise}.
  \end{cases}
\]
If only a ground-truth point \(\mathbf{p}^*\) is available, we use a distance threshold \(\theta\):
\[
  r_{c}(\mathbf{p}, \mathbf{p}^*) =
  \begin{cases}
    1, & \|\mathbf{p}-\mathbf{p}^*\|_2 \le \theta,\\
    0, & \text{otherwise}.
  \end{cases}
\]

In practice, we sum the applicable components to form an outcome reward: $r_o = r_{d} + r_{c}$.
This simple yet flexible scheme can cover classification, numeric reasoning, and grounding.
See Appendix~\ref{app sec: rewards} for details on adapting these rewards alongside the format reward to individual downstream tasks.

%% file: sec/exps.tex
\section{Experiments}
\label{sec: experiments}
In this section, we conduct experiments on various benchmarks to evaluate our approach. Mainly, we design the experiments to study the following key questions:

\noindent{$\mathbf{Q1}$:} Compared to vanilla GRPO, how does \ours impact performance and efficiency?

\noindent{$\mathbf{Q2}$:} Is there a correlation between \ours's skipping behavior and the strength of reasoning ability (\eg~different model sizes or a single model under different iterations)?

\noindent{$\mathbf{Q3}$:} Do we really need SFT with thought dropout? Can we rely solely on prompt following if the base model is strong enough?

\subsection{Benchmarks and Settings}
To evaluate the effectiveness and generalization ability of our approach on the below settings:

\begin{wraptable}{r}{0.6\textwidth}
\vspace{-10pt}
\centering
\scriptsize
\setlength{\tabcolsep}{1.5pt}
\captionsetup{font=small}
\caption{\textbf{Summary of benchmark used in our evaluation}.}
\begin{tabular}{lcccccc}
\toprule
\textbf{Benchmark} & \textbf{Model} & \textbf{OOD} & \textbf{Type} & \textbf{Difficulty} & \textbf{Answer} & \textbf{Thought len.} \\
\midrule
GSM8K~\cite{cobbe2021training} & LLM & & Math & Hard & Number & $939$ \\
CLEVR~\cite{johnson2017clevr} & VLM & & Counting & Easy & Integrate & $586$ \\
Super-CLEVR~\cite{li2023super} &VLM & \cmark &  Counting & Easy & Integrate & -- \\
GeoQA~\cite{chen2021geoqa} & VLM & & Math & Hard & Number & $1652$\\
AITZ~\cite{aitz} & Agent & & GUI & Medium & Action (x)  & $283$\\
AITZ (OOD) & Agent & \cmark & GUI & Medium & Action (x) & $283$\\
\bottomrule
\end{tabular}
\label{tab:dataset_summary}
\vspace{-10pt}
\end{wraptable}

\noindent\textbf{Benchmarks.}
We evaluate \ours on three vision-language benchmarks, including the general benchmark CLEVR~\cite{johnson2017clevr} (3D object counting), agent benchmark AITZ~\cite{aitz} (mobile navigation), and the math benchmark GeoQA~\cite{chen2021geoqa} (middle school math questions) as illustrated in Table~\ref{tab:dataset_summary}, spanning a spectrum of reasoning levels from simple to complex. 
To benchmark the model's Out-of-Distribution (OOD) performance, we  also evaluate on Super-CLEVR~\cite{li2023super} to supplement the CLEVR.
AITZ comprises four test domains: we train on the \{General\} and test on the remaining OOD domains: \{Web shopping, Google apps, Install\}.
We remove the choices in GeoQA and ask the model to generate the answer, enhancing the reasoning complexity.
AITZ includes action thought annotations, which we utilize directly, while applying our reverse thinking to generate thoughts for SFT on CLEVR and GeoQA. More benchmark details refer to Appendix~\ref{app sec: dataset}.

\noindent\textbf{Training details.}
We conduct our experiments using {Qwen-2.5-VL-Instruct-3B/7B}~\cite{bai2025qwen2} as the base model. 
All experiments are conducted utilizing 8 NVIDIA H20 GPUs. 
We train $100$ steps for both CLEVR and AITZ, and $300$ epochs for GeoQA, given its higher reasoning difficulty level.
See setup details in Appendix~\ref{app sec: setup}.
We leverage vLLM~\cite{vllm} to accelerate GRPO training.
We add the SFT stage before GRPO as the baseline on the agent task with the same setting as \ours because we observe that directly applying GRPO would cause the 0 coordinate reward during the training process, considering its complex output format.
For simplicity, we set the dropout probabilities to 50\% and examine the impact of different dropout ratios selected from $\{20\%, 50\%, 80\% \}$ in Sec~\ref{subsec: skip ratio, droup ratio}.

For evaluation, we test all the datasets under the greedy strategy.
In CLEVR and GeoQA tasks, where answers are numerical, we measure accuracy by comparing the predicted number to the ground truth. 
In the AITZ task, where answers are structured as JSON-formatted actions, we report step-level and task-level metrics, including type accuracy (correct action type) and exact accuracy (correct action type and click coordinates) following~\cite{showui}. 

\begin{table*}[t]
\centering
\captionsetup{font=small}
\caption{
\textbf{Performance comparison between \ours and vanilla GRPO.}
Acc. is the accuracy on the test set. 
Time is the RL training time.
Length is the average competition length at the end of training.
}
\resizebox{\textwidth}{!}{
\begin{tabular}{llllllllll}
\toprule
& \multicolumn{3}{c}{\textbf{Counting (CLEVR)}}  & \multicolumn{6}{c}{\textbf{Math (GeoQA)}} \\
\cmidrule(lr){2-4} \cmidrule(lr){5-10}
& {Acc} ($\uparrow$) & {Time } ($\downarrow$)& {Len.} ($\downarrow$) & {Acc} ($\uparrow$) & {Time } ($\downarrow$)& {Len.} ($\downarrow$) & {Acc} ($\uparrow$) & {Time } ($\downarrow$)& {Len.} ($\downarrow$)\\
& \multicolumn{3}{c}{Qwen-2.5-VL-3B} & \multicolumn{3}{c}{Qwen-2.5-VL-3B} & \multicolumn{3}{c}{Qwen-2.5-VL-7B}\\
\midrule
\quad Baseline & $64.0$ & - & $306$  & $36$ & - & $924$ &  $38$ & - & $939$\\
\quad w. GRPO & $93.5$ & $1$h $44$m & $227$  & $37$ & $2$h $50$m & $272$ & $44$ & $3$h $4$m & $1069$\\
\quad w. \ours  & $98.5$ & $57$m   & $28$\textcolor{red}  & $51$ & $2$h $4$m  & $96$  & $61$ & $2$h $34$m & $112$ \\
\midrule
\quad \textbf{Gain} & \textcolor{red}{$+5.0$} & \textcolor{red}{$-47$m} & \textcolor{red}{$-87\%$} & \textcolor{red}{$+14$} & \textcolor{red}{$-46$m} & \textcolor{red}{$-65\%$} & \textcolor{red}{$+17$} & \textcolor{red}{$-29$m} & \textcolor{red}{$-90\%$} \\
\bottomrule
\end{tabular}
}
\label{tab: clever math results}
\vspace{-2mm}
\end{table*}
\begin{figure}[t]
    \centering

    \begin{subfigure}[b]{0.24\textwidth}
        \includegraphics[width=\textwidth]{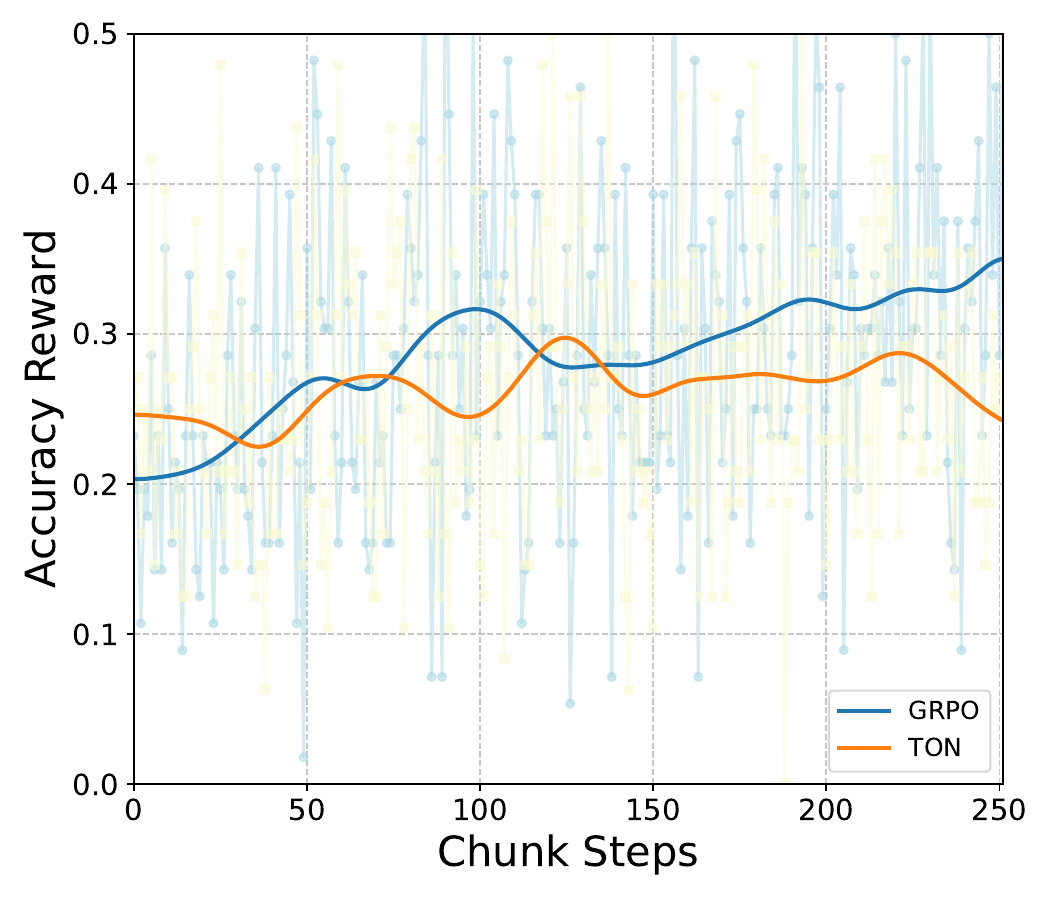} 
        \caption{Training rewards}
        \label{fig:math reward}
    \end{subfigure}
     \hfill
     \begin{subfigure}[b]{0.24\textwidth}
        \includegraphics[width=\textwidth]{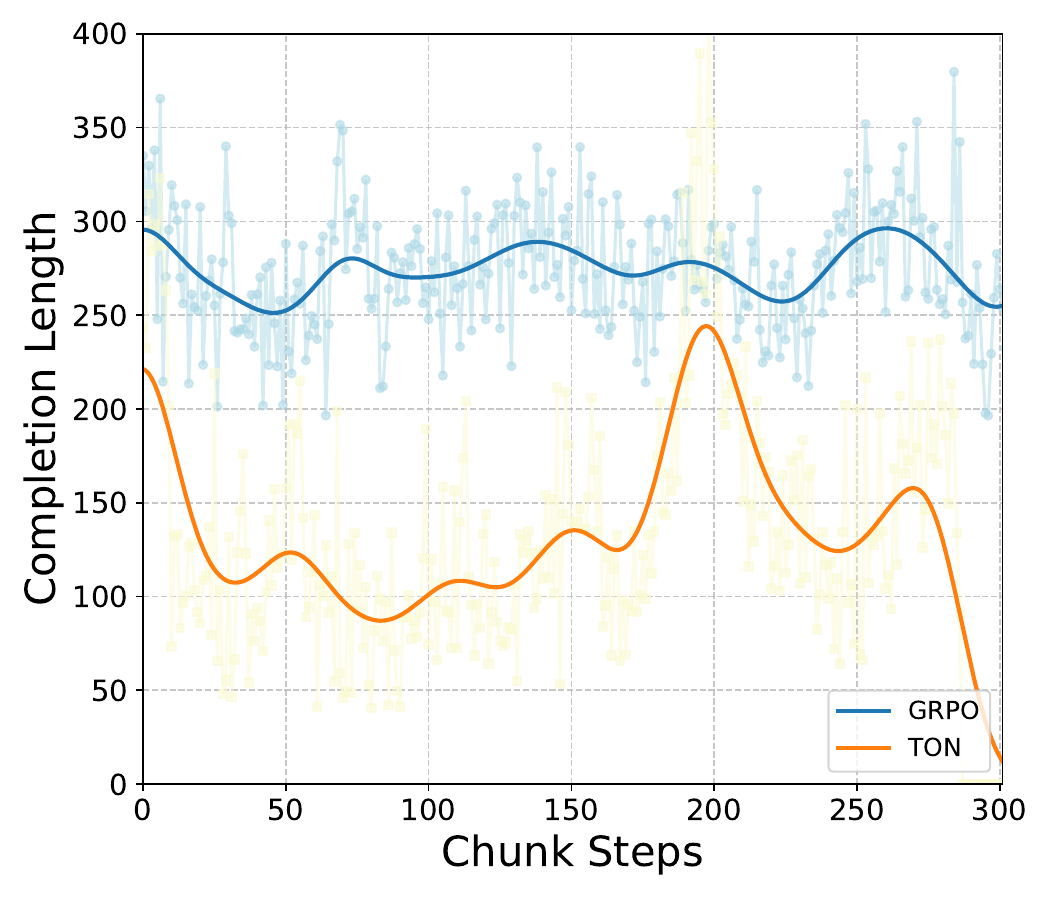} 
        \caption{Completion length}
        \label{fig: math length}
    \end{subfigure}
    \hfill
     \begin{subfigure}[b]{0.24\textwidth}
        \includegraphics[width=\textwidth]{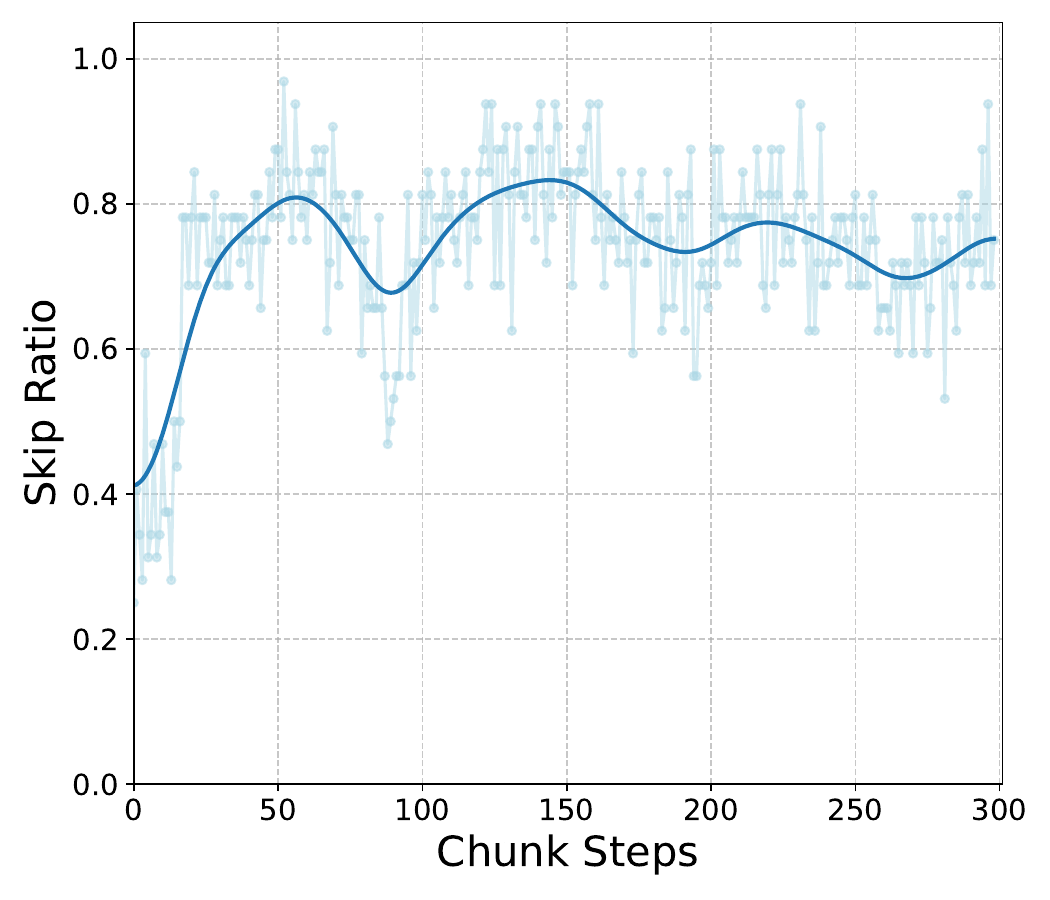} 
        \caption{Skip think ratio
        }
        \label{fig: math skip}
    \end{subfigure}
    \hfill
     \begin{subfigure}[b]{0.24\textwidth}
        \includegraphics[width=\textwidth]{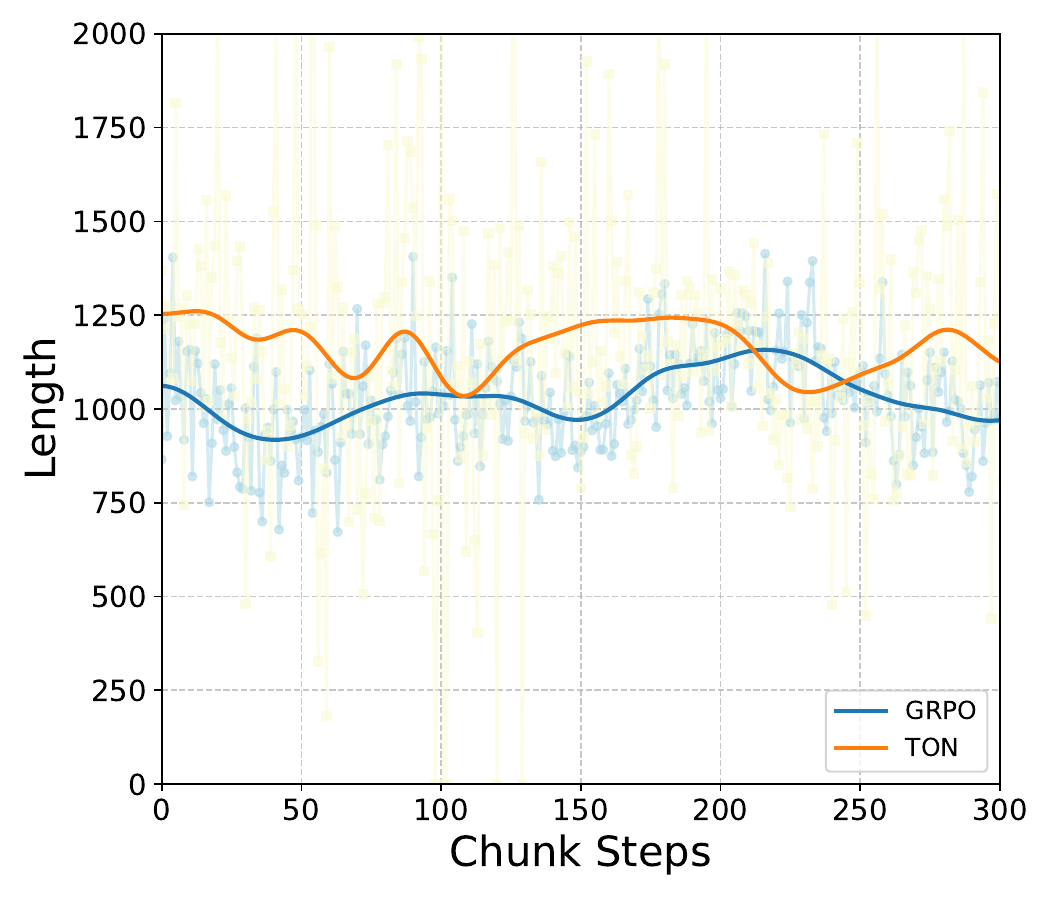} 
        \caption{Output len. with think}
        \label{fig: math whole length}
    \end{subfigure}
   
    
\captionsetup{font=small}
\caption{\textbf{Training metrics comparison between \ours and GRPO on GeoQA.} 
(a) Training rewards, 
(b) Completion length over training steps, 
(c) Ratio of non-think outputs to total samples at each step for \ours, and 
(d) Average completion length of think outputs across training.}
    \label{fig: math metric all}
    \vspace{-2mm}
\end{figure}

\subsection{$\mathbf{Q1}$: Performance and Efficiency Comparison between \ours and GRPO}
\label{subsec: performance}

In Table~\ref{tab: clever math results}, we present \ours on the CLEVR and GeoQA benchmarks under both 3B and 7B settings, with the performance, time consumption, and the average completion length at the RL stage.
We find that \ours effectively reduces the average of the completion length by up to $90\%$ while achieving comparable even superior performance compared to GRPO with a maximum of $17$ Acc. gains. This imply that 
\textbf{skipping unnecessary reasoning can lead to better performance.}
The reduction of the completion length decreases the decoding time when generating samples, thus simultaneously shortening the training time.
Figure~\ref{fig:math reward} \& \ref{fig: math length} show the reward and completion length curves where \ours remains the rewards on par with vanilla GRPO while the completion length reduces significantly.
Appendix~\ref{app: ton on general} \& \ref{app: ton on math} shows the entire metrics during training.

\begin{table*}[t]
\centering
\captionsetup{font=small}
\caption{ \textbf{Out-of-domain (OOD) performance comparison} between our method \ours and GRPO on the the AITZ -- multi-step mobile navigation.
‘Type’ is the action type accuracy and ‘Exact’ requires both the type and value to be correct exactly.
‘Avg.’ is the average accuracy of all domains.
‘Task-level thought's’ is the average output lengths on all OOD domains.
Step-level accuracy is reported.
}
\resizebox{\textwidth}{!}{
\begin{tabular}{llllllllllllll}
\toprule
& \multirow{3}{*}{\textbf{Think?}}
& \multicolumn{2}{c}{\textbf{IID}}  
& \multicolumn{6}{c}{\textbf{OOD}} 
& \multicolumn{2}{c}{\multirow{2}{*}{\textbf{Avg}}} 
& \multirow{2}{*}{\textbf{Task-level}} \\
\cmidrule(lr){3-4} \cmidrule(lr){5-10} 
& & \multicolumn{2}{c}{General} & \multicolumn{2}{c}{Google apps} & \multicolumn{2}{c}{Web} & \multicolumn{2}{c}{Install}  & & & \\
&  & {type} & {exact} & {type} & {exact} & {type} & {exact} & {type} & {exact} & {type} & {exact} & \textbf{Thought's len.} \\
\midrule
Qwen-2.5-VL-3B & \cmark & $0.01$ & $0$ & $0.01$ & $0$ & $0.01$ & $0$ & $0.01$ & $0$ & $0.01$ & $0$ & $2132$\\
\quad~w. SFT & \xmark & $0.39$ & $0.11$ & $0.44$ & $0.12$ & $0.54$ & $0.19$ & $0.47$ & $0.17$ & $0.46$ & $0.15$ & $742$\\
\quad~w. SFT & \cmark & $0.67$ & $0.12$ & $0.53$ & $0.17$ & $0.56$ & $0.13$ & $0.58$ & $0.14$ & $0.58$ & $0.14$ & $3572$\\
\quad~w. GPRO & \cmark & $0.74$ & $0.6$ & $0.72$ & $0.57$ & $0.7$ & $0.5$ & $0.81$ & $0.65$ & $0.74$ & $0.59$ & $3664$ \\
\quad~w. \ours & Ours & $0.74$ & $0.6$ & $0.74$ & $0.56$ & $0.72$ & $0.5$ & $0.78$ & $0.64$ & $0.75$ & $0.59$ & $922$\\
\quad\quad \textcolor{gray}{\textit{Gain}} &  & \textcolor{gray}{+$0.0$} & \textcolor{gray}{+$0.0$} & \textcolor{red}{+$0.02$} & \textcolor{gray}{-$0.01$} & \textcolor{red}{+$0.02$} & \textcolor{gray}{+$0.0$} & \textcolor{gray}{-$0.03$} & \textcolor{gray}{-$0.01$} & \textcolor{gray}{+$0.01$} & \textcolor{gray}{+$0.0$} & \textcolor{red}{\textbf{$-2742$}} \\
\midrule
Qwen-2.5-VL-7B & & $0.28$ & $0.14$ & $0.26$ & $0.1$ & $0.33$ & $0.13$ & $0.39$ & $0.16$ & $0.31$ & $0.13$ & $3304$\\
\quad~w. GRPO & \cmark & $0.64$ & $0.22$ & $0.73$ & $0.32$ & $0.6$ & $0.15$ & $0.62$ & $0.23$ & $0.65$ & $0.23$ & $3272$\\
\quad~w. \ours & Ours & $0.74$ & $0.54$ & $0.62$ & $0.23$ & $0.68$ & $0.47$ & $0.73$ & $0.55$ & $0.69$ & $0.45$ & $908$\\
\quad\quad \textcolor{gray}{\textit{Gain}} &  & \textcolor{red}{+$0.1$} & \textcolor{red}{\textbf{+$0.32$}} & \textcolor{gray}{-$0.11$} & \textcolor{gray}{-$0.09$} & \textcolor{red}{+$0.08$} & \textcolor{red}{\textbf{+$0.32$}} & \textcolor{red}{+$0.09$} & \textcolor{red}{\textbf{+$0.32$}} & \textcolor{red}{+$0.04$} &  \textcolor{red}{\textbf{+$0.22$}} & \textcolor{red}{\textbf{$-2364$}} \\
\bottomrule
\end{tabular}
}
\label{tab:aitz}
\vspace{-2mm}
\end{table*}

\begin{figure}[t]
\centering

\hfill
\begin{subfigure}[b]{0.24\textwidth}
    \includegraphics[width=\linewidth]{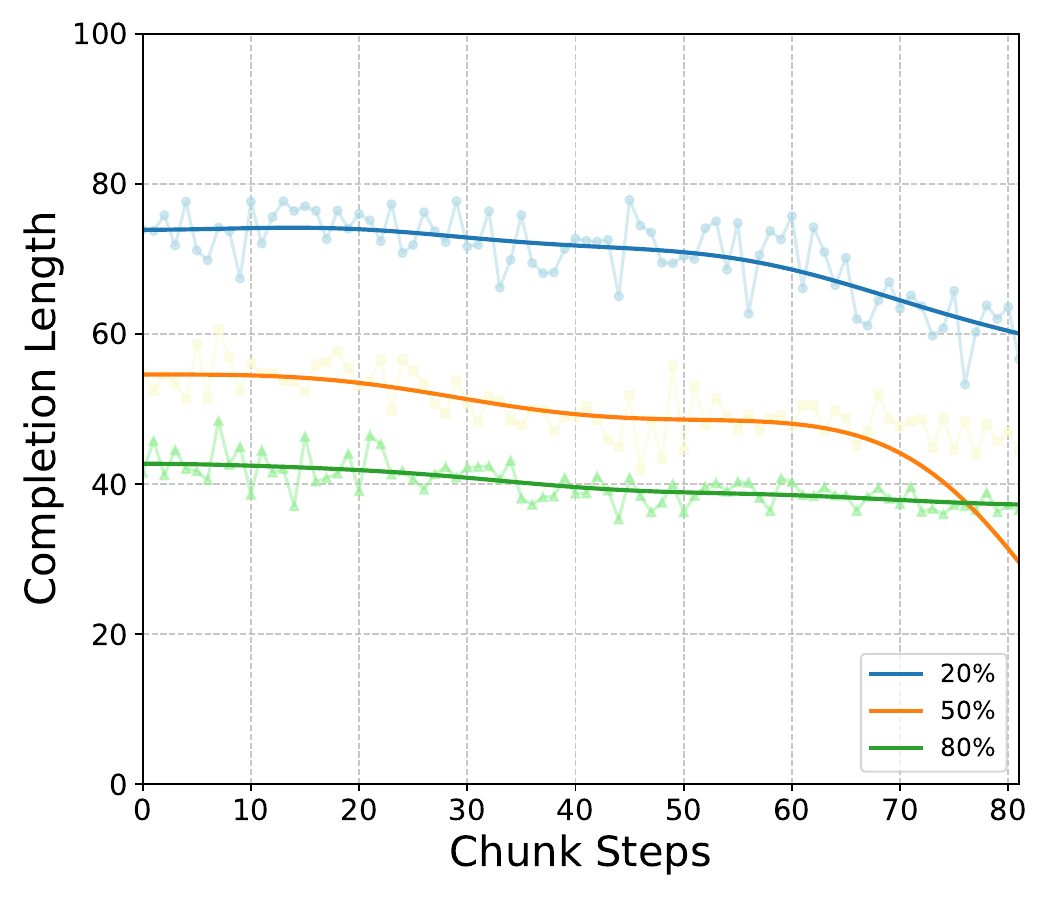}
    \caption{Completion length
    }
    \label{fig: mask ratio length}
\end{subfigure}
\hfill
\begin{subfigure}[b]{0.24\textwidth}
    \includegraphics[width=\linewidth]{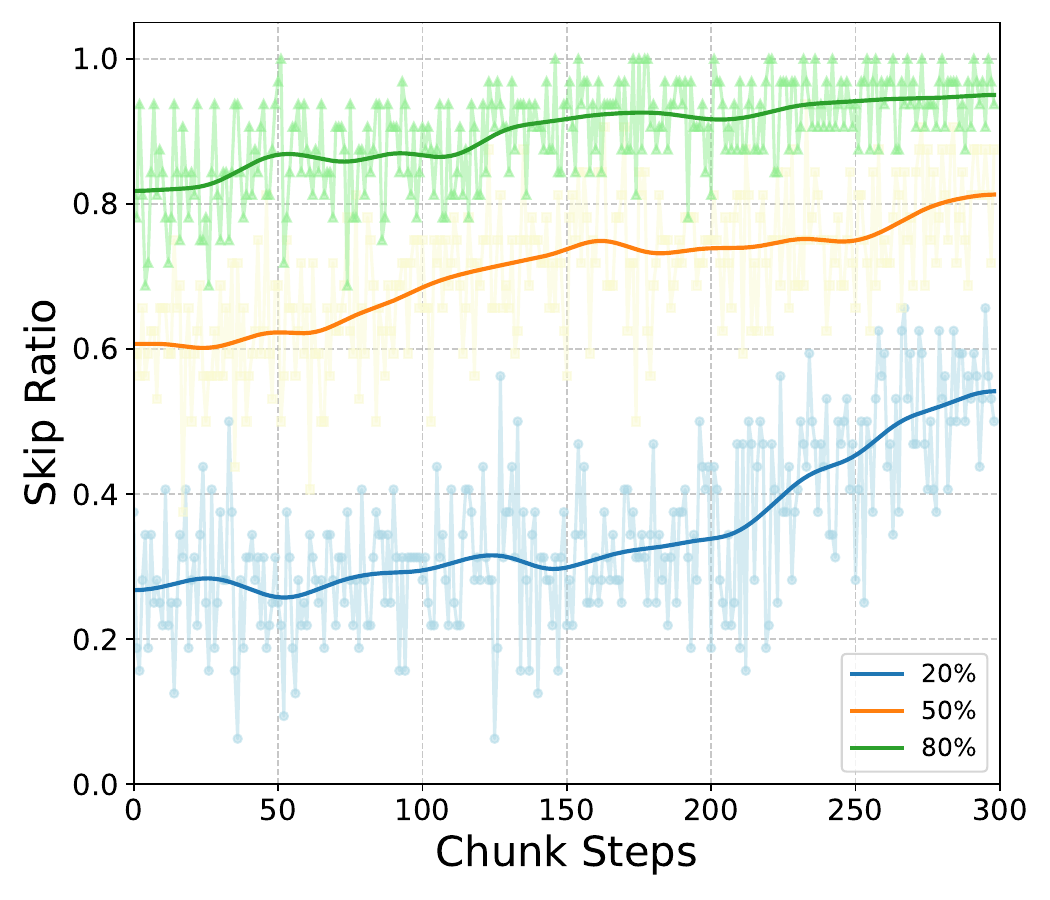}
    \caption{Skip think ratio}
    \label{fig: mask ratio skip}
\end{subfigure}
\hfill
\begin{subfigure}[b]{0.24\textwidth}
    \includegraphics[width=\linewidth]{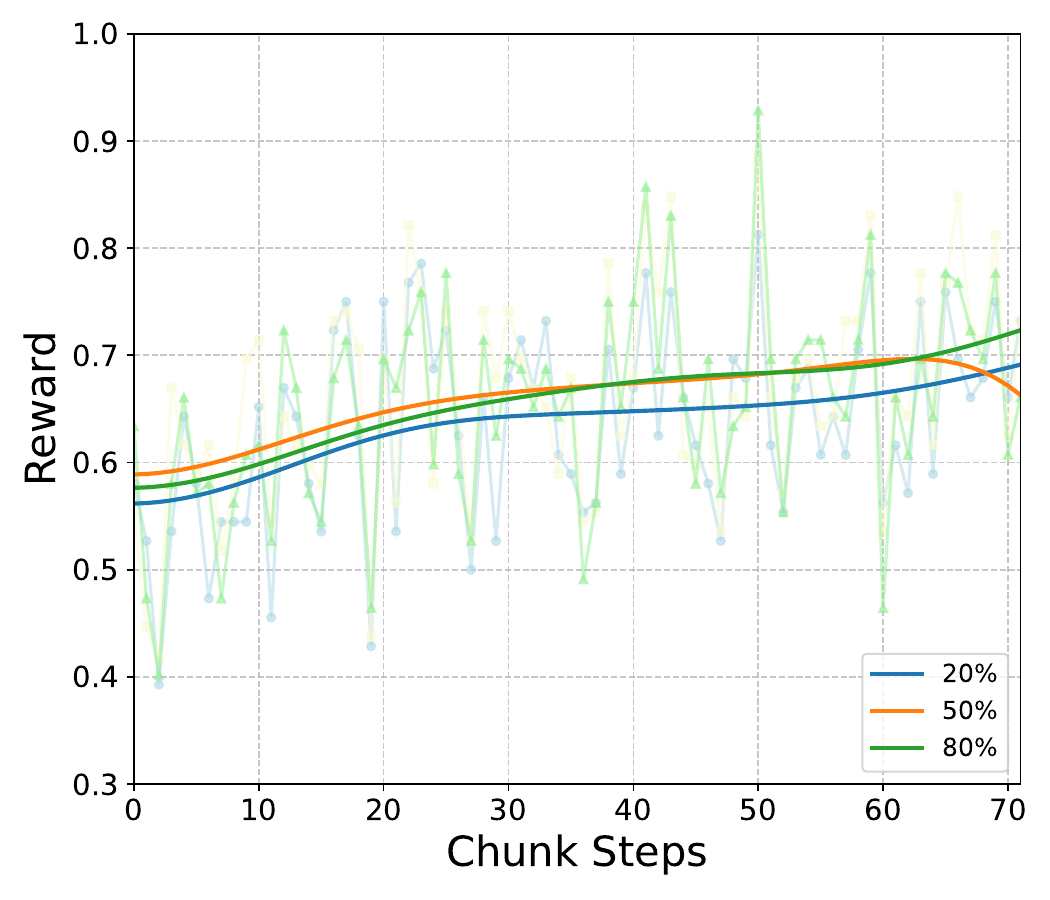}
    \caption{Training rewards}
    \label{fig: mask ratio reward}
\end{subfigure}
\begin{subfigure}[b]{0.24\textwidth}
    \includegraphics[width=\linewidth]{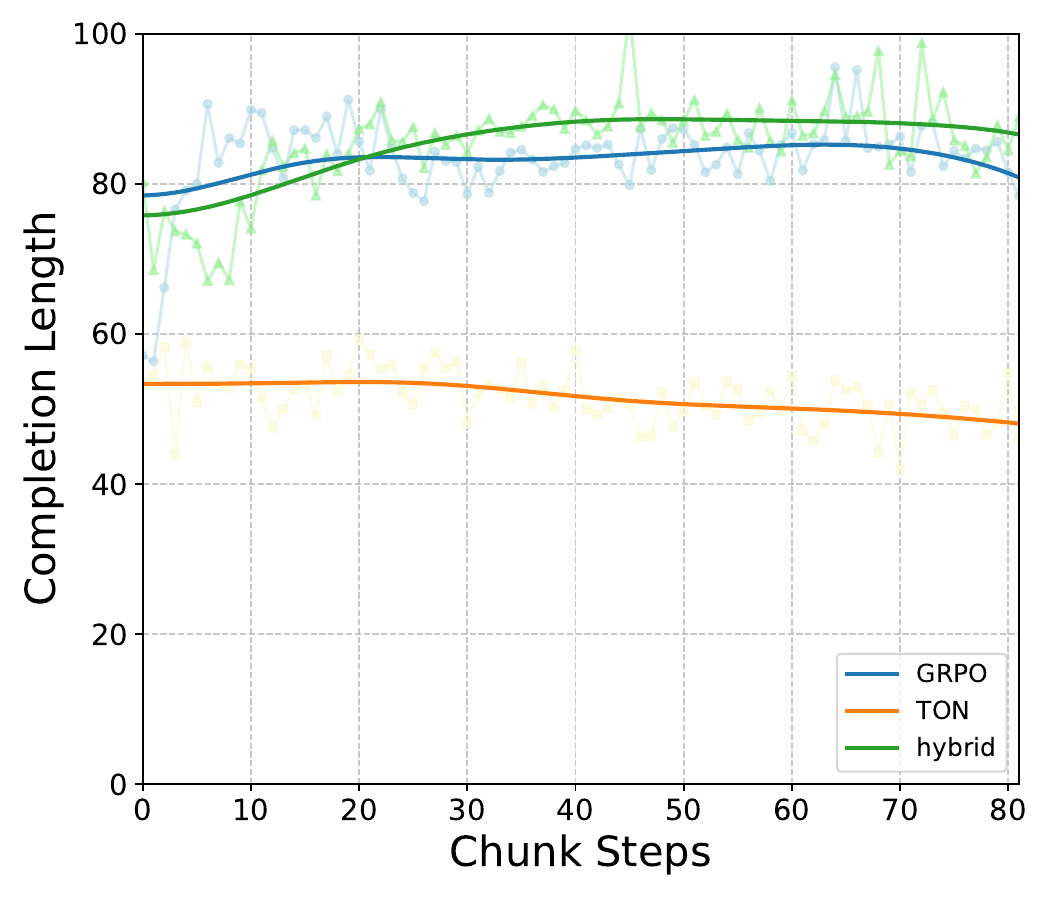}
    \caption{ Prompting v.s. SFT
     }
    \label{fig: hybrid}
\end{subfigure}
\captionsetup{font=small}
\caption{\textbf{Further Analysis of \ours on the AITZ benchmark.} (a)(b)(c) is the average completion length, skip thought ratios, and the reward under different dropout probabilities. (d) Prompting (hybrid) does not reduce the completion length, while \ours using SFT can effectively reduce it.
}
\label{}
\end{figure}
\begin{figure}
    \centering
    \includegraphics[width=\linewidth]{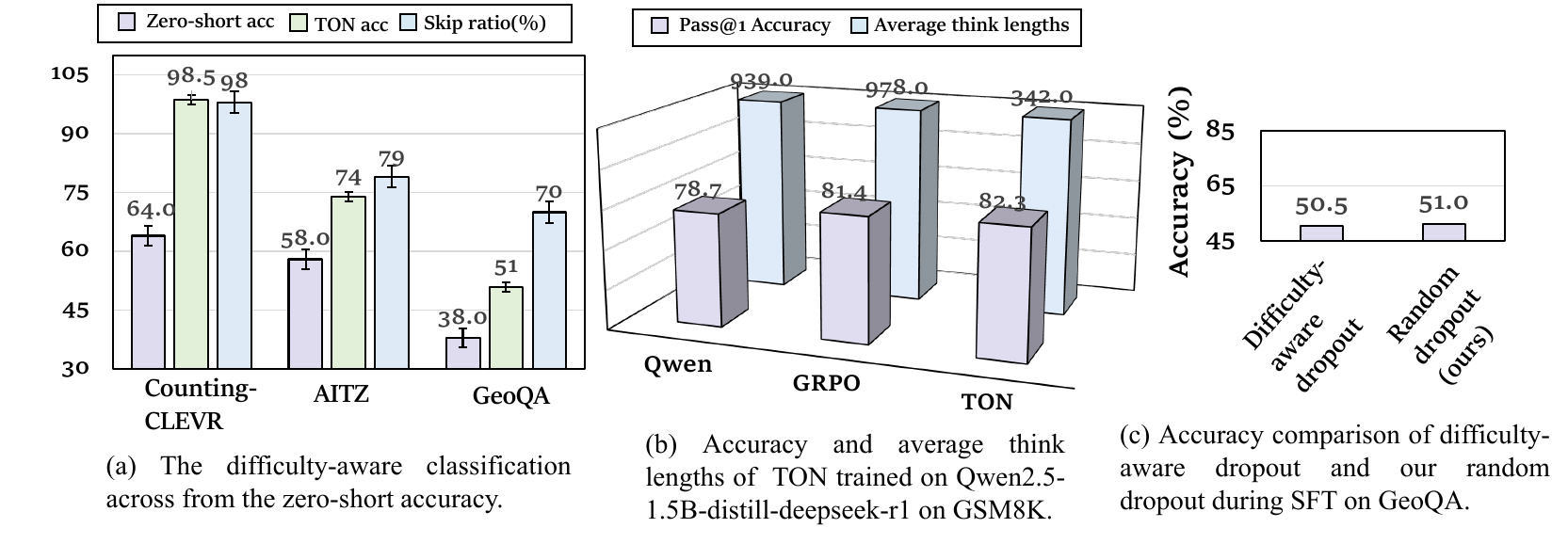}
    \caption{\textbf{Further Analysis of \ours.} (a)(c) give the in-depth analysis of the effectiveness of our thought dropout and the generalization across the task difficulty. (b) shows the superiority of \ours on the text-only domains.}
    \label{fig:rebuttal}
\end{figure}

\textbf{Multi-step Navigation and OOD Testing.}
In Table~\ref{tab:aitz}, we evaluate \ours's performance on AITZ -- multi-step mobile navigation, we also assessed its generalization capabilities on OOD test sets using a greedy decoding strategy. 
Table~\ref{tab:aitz} summarizes the step-level type match accuracy and exact match accuracy for both IID (general) and OOD (Google Apps, web shopping, and install) domains on AITZ, with detailed training visualization in Appendix~\ref{app: ton on agent}. 
Overall, \ours\ demonstrates comparable OOD generalization performance to GRPO, while significantly reducing the task-level output length from $3K$ to $0.9K$ ($\mathbf{70\%}$ \textbf{token saving}). 
This highlights the strong potential of \textbf{TON to substantially reduce completion length without compromising performance}.
See Appendix~\ref{app sub: ood performance on general} for the OOD performance on other  benchmarks. 

\textbf{Adapt \ours framework to the text-only setting.} Furthermore, we extend our study to the LLM domain and present the corresponding experiments and results in Figure~\ref{fig:rebuttal}(b) on the LLM benchmark GSM8K. The findings indicate that TON significantly reduces response length while maintaining high accuracy, demonstrating the generalizability of TON across different modalities.

\subsection{$\mathbf{Q2}$: Skip Thought Ratio Analysis}
\label{subsec: skip ratio, droup ratio}
Beyond the performance change and completion length reduction achieved by \ours, we further investigated the evolution of the skip ratio in ‘Thought dropout’ during the training step.
Figure~\ref{fig: math skip} illustrates the percentage of skip ratio in the generated samples at each step on GeoQA. 
We observed an increasing trend in the skip ratio during the training process with the increase in training reward.
A similar trend is observed across three benchmarks in Figure~\ref{app fig:  skip ratio} in the Appendix~\ref{app sec: skip thought ratio}. This phenomenon suggests that \textbf{the model progressively internalizes the reasoning process—learning to skip explicit thoughts while still producing accurate answers.}
Moreover, Figure~\ref{fig: math whole length} illustrates the length of these outputs generated with ‘think’. 
\ours~maintain comparable lengths to the vanilla GRPO, indicating that the \ours model can \textbf{choose not to think but remains diligent when deeper reasoning is necessary.}

\textbf{Thought dropout ratio ablation.}
We experiment with the impact of different thought dropout ratios of $20\%$, $50\%$, and $80\%$ during the SFT stage. 
Figure~\ref{fig: mask ratio length} \& \ref {fig: mask ratio skip} show the completion lengths and the skip ratio during the training process on AITZ.
Figure~\ref{fig: mask ratio reward} shows a close reward curve of these three variants.
Refer more metrics on Appendix~\ref{app sec: different dropout ratio}.
Although the dropout ratios differ, \ours consistently exhibits an increasing skip ratio as training progresses. Notably, the $20\%$ setting shows a rapid increase in skip rate, while the higher $80\%$ setting remains relatively stable throughout training.
This motivates us to start with a lower dropout probability for further investigation. TON can then be dynamically optimized according to reward signals—decreasing the dropout ratio when performance is high and increasing it when performance drops.


\textbf{Deep analysis of the difficulty-aware dropout and random dropout (ours).}
As shown in Figure~\ref{fig:rebuttal}(a), we further assess the task difficulty both qualitatively (via intuitive reasoning complexity) and quantitatively (by base VLM zero-shot accuracy), grouped as: 60-100 (easy), 40-60 (medium), and 0-40 (hard).
Our findings indicate that our TON maintains a high skip ratio for easy counting queries, allowing it to bypass unnecessary thinking processes, while keeping a low skip ratio for harder, more professional, knowledge-intensive questions to fully utilize its thinking capabilities.
Furthermore, we do the difficulty-aware ablation as shown in Figure~\ref{fig:rebuttal}(c), where we drop thoughts for only the easy samples (those answered correctly by the base VLM). 
To investigate this, we then implement a difficulty-aware dropout strategy and compare its accuracy under our TON training framework where random dropout slightly outperforms difficulty-aware dropout (51.0\% vs. 50.5\% in TON Accuracy).
The results suggest that hand-crafted heuristics for task difficulty may introduce noise or unintended bias, potentially interfering with the learning process while our random dropout offers a simpler, unbiased alternative that generalizes well across tasks.

\subsection{$\mathbf{Q3}$: Emprical Verfication of SFT Significance in \ours}
\label{subsec: prompt vs sft}

In addition to incorporating the skip-think format during the SFT stage as in \ours, we explored a simpler alternative: modifying the prompt to encourage the model to automatically omit reasoning steps, enabling direct GRPO training without the need for a separate SFT stage.
The hybrid-thought prompt is defined as follows:
\begin{AIbox}{Prompt for \textit{Hybrid Thinking}}
    A conversation between User and Assistant. The user asks a question, and the Assistant solves it. The assistant evaluates whether the question is simple enough to answer directly. \\
    \textcolor{blue}{If simple, the output is formatted as  <think>\textbackslash n\textbackslash n<\textbackslash think><answer>answer here<\textbackslash answer>.} \\
    If the question is difficult, the assistant needs to first think then answering the question. 
    The output is formatted as <think> reasoning process here <\textbackslash think><answer> answer here <\textbackslash answer>.\\
    The assistant is encouraged to use <think>\textbackslash n\textbackslash n<\textbackslash think> while maintaining accuracy.
\end{AIbox} 

Figure~\ref{fig: hybrid} shows the completion length of GRPO using the hybrid prompt, vanilla GRPO (with a full-think prompt), and \ours throughout the training process on AITZ.
Appendix~\ref{app sec: prompt sft} presents similar trends across three benchmarks, revealing only minor differences in completion length between the hybrid prompt and vanilla GRPO.
Moreover, we observe only 2 ‘skip’ cases in GeoQA and none in AITZ among all samples generated by GRPO during both training and inference.
We attribute this to the model’s tendency to play it safe by generating long and detailed reasoning, consistent with its ingrained behavioral patterns learned during pre-training or SFT.
Since the model does not produce skip-thought outputs, applying additional reward to these outputs has no effect, resulting in a zero contribution throughout training.
These findings highlight the necessity of our SFT stage with thought dropout (Sec.~\ref{sec: first sft stage}) to establish the desired format-following behavior.
\begin{figure}[t]
    \centering
    \includegraphics[width=0.9\linewidth]{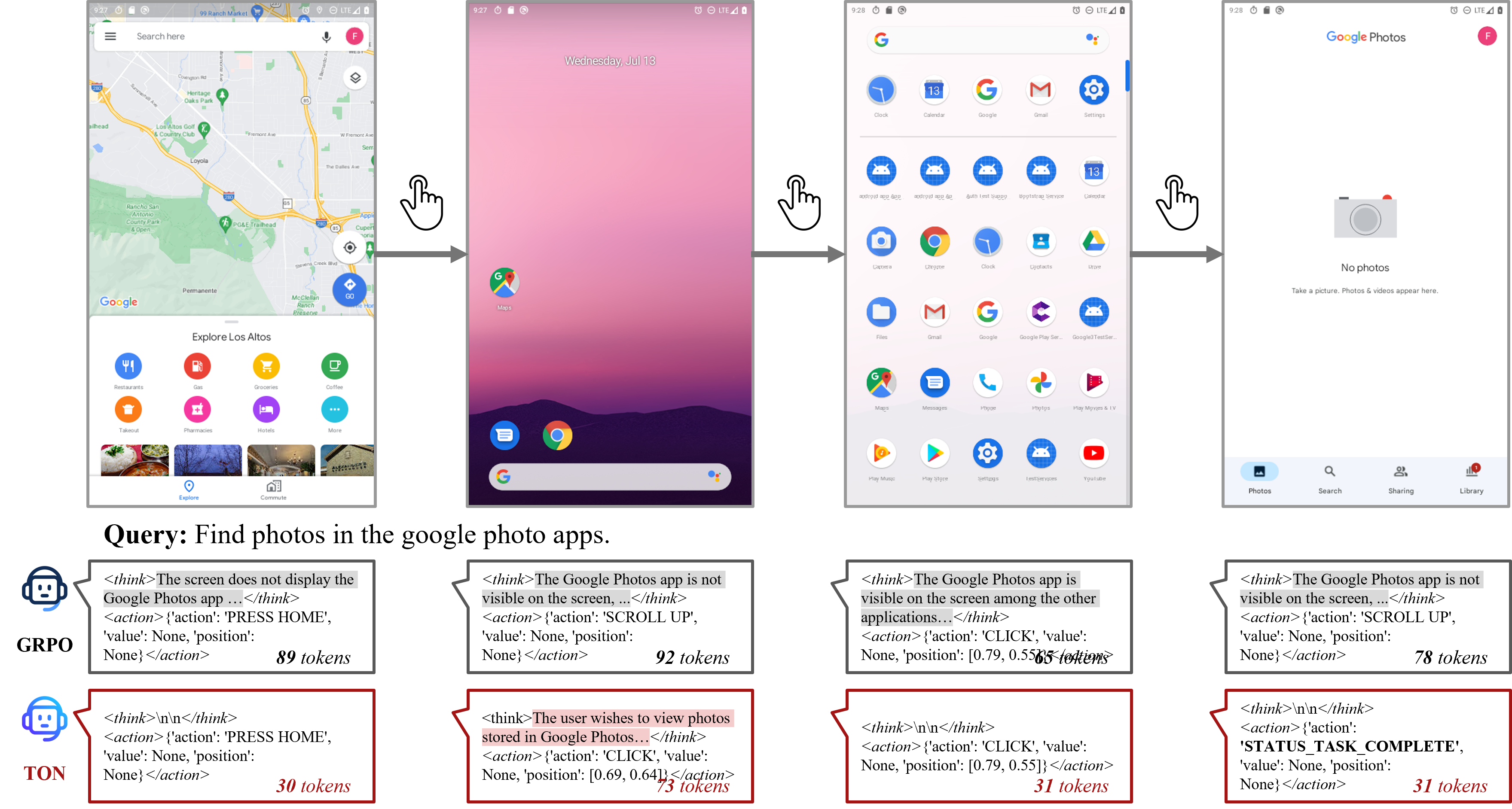}
    \captionsetup{font=small}
    \caption{\textbf{Comparison between GRPO and TON on Agent setting AITZ~\cite{aitz}}.
    TON adaptively skips unnecessary think steps during multi-step mobile navigation, achieving greater decoding efficiency compared to GRPO while maintaining task accuracy (saving 60\% tokens in this case).
    }
    \vspace{-3mm}
    \label{fig:aitz}
\end{figure}
\begin{table*}
  \begin{minipage}{0.99\textwidth}
\centering
\captionsetup{font=small}
\captionof{table}{
\textbf{Illustration between Thinking and Non-Thinking modes from CLEVR}~\cite{johnson2017clevr}. 
TON demonstrates selective activation of reasoning—engaging thought only when needed—whereas GRPO generates reasoning traces for both cases indiscriminately.
The full outputs are in Table~\ref{tab:clevr example}.
}

\scalebox{0.8}{
\begin{tabular}{l p{5.4cm} p{6.0cm} }
\toprule
 \multicolumn{3}{l}{\bf Counting examples from CLEVR~\cite{johnson2017clevr}}  \\
\midrule
&  \includegraphics[height=3.5cm]{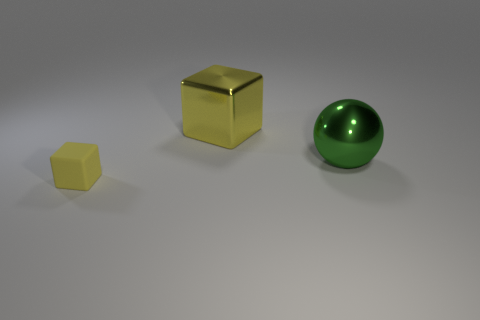} 
&  \includegraphics[height=3.5cm]{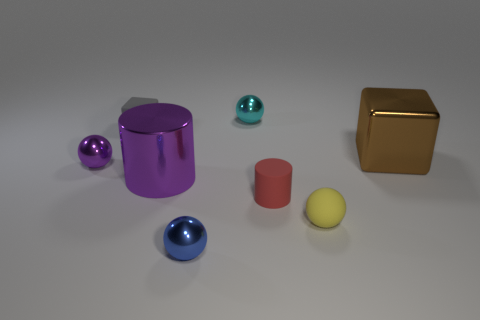} \\
\midrule
Question (\textcolor{gray}{GT}) & How many items are there in the image? \textcolor{gray}{{<answer>3</answer>}} & How many items are there in the image? \textcolor{gray}{{<answer>8</answer>}}
\\ 
\midrule
GRPO's outputs & 
<think>
The image shows three objects ...

1. Green Cylinder: Transparent (metallic)
2. Teal Sphere: Transparent (metallic)

</think>

<answer>3</answer>
& 
<think>
- I count six distinct objects in the picture: a large green matte cylinder, a blue matte cylinder, 
...

- I also note the size and color variety of each object as described by the visual cues.
</think>

<answer>10</answer>
\\ \midrule
TON's outputs & <think>\textbackslash n\textbackslash n</think>

<answer>3</answer> & 
<think>
Okay, let's check this out. The user provided a image with
....

First, I need to count ...

1. "A small purple rubber sphere..."... 
\textcolor{blue}{3. "A small gray rubber cube..." Third.}

...



Wait, that's eight distinct entries...
</think>

<answer> 8 </answer>
\\ \bottomrule
\end{tabular}
}

\label{main tab: clevr example}  
  \end{minipage}
\end{table*}

\subsection{Qualitative Examples}
Figure~\ref{fig:aitz} compares GRPO and TON on the AITZ benchmark for multi-step mobile navigation. While GRPO generates verbose reasoning at every step, TON adaptively skips unnecessary thinking, reducing token usage by 60\% without sacrificing task accuracy. This demonstrates TON’s efficiency in handling real-world, \textit{long-horizon procedural agent tasks}.
Table~\ref{main tab: clevr example} further illustrates TON’s ability to selectively activate reasoning. Unlike GRPO, which consistently generates detailed thought traces, TON omits reasoning for simple questions that can be answered at a glance, while producing accurate and focused reasoning for complex scenarios involving visual occlusion.

%% file: sec/conclusion.tex
\section{Conclusion}

We present \ours, a simple yet effective two-stage training framework that enables vision-language models to learn \textit{when to reason}—introducing selective reasoning as a controllable and trainable behavior. By combining thought dropout during supervised fine-tuning with reward-guided refinement via GRPO, \ours significantly reduces completion length (up to $\mathbf{90\%}$) without sacrificing—and in some cases improving—performance across diverse reasoning tasks. 
Our findings challenge the assumption that full reasoning traces are always beneficial and pave the way for more efficient, human-like reasoning strategies in both multimodal intelligence and reinforcement learning.

%% file: sec/ack.tex
\section{Acknowledgements}

This research is supported by the National Research Foundation, Singapore under its AI Singapore Programme (AISG Award No: AISG3-RP-2022-030).

We sincerely thank Dongchi Huang for his invaluable guidance on the code and for providing essential computational resources. 
We also appreciate Binghui Xie's insightful discussion on topic selection and idea suggestions. 
Additionally, we are grateful to Qiguang Chen and Yuxuan Wan for their thoughtful and constructive feedback on this paper.
Finally, we extend our gratitude to Xiaojun Guo and Qixun Wang for their valuable advice on visual reasoning and the GRPO series methods.

%% file: sec/appendix.tex
\begin{center}
	\LARGE \bf {Appendix}
\end{center}

\etocdepthtag.toc{mtappendix}
\etocsettagdepth{mtchapter}{none}
\etocsettagdepth{mtappendix}{subsubsection}
\tableofcontents
\newpage


\section{Motivation Experiments}
\label{app sec: motivation exps}

Table~\ref{app tab: motivation} presents the performance of the agent VQA with and without the think source during the SFT stage, as well as with and without the think format in the GRPO reward function.
Using the think source results in higher performance but longer output lengths, while excluding it leads to lower performance with shorter outputs.

\begin{table*}[h!]
    \centering
  \caption{Qwen2.5-VL-3B on the agent dataset (Android-in-the-zoo) with/ without think before the answer by SFT and vanilla GRPO. acc is the test accuracy and len is the output length of step-level. 
  }
    \resizebox{0.5\linewidth}{!}{
  \begin{tabular}{ccccc}
    \toprule
     & \multicolumn{2}{c}{SFT}  & \multicolumn{2}{c}{GRPO} \\
     & acc & len  & acc & len \\
    \midrule
    w think & 0.67  & 181 & 0.74 & 88\\
    wo think & 0.39 & 29 & 0.58 & 29\\
    \bottomrule
  \end{tabular}
  }
  \label{app tab: motivation}
\end{table*}

\section{Rewards for Downstream Tasks}
\label{app sec: rewards}
General VQA: $r=r_f+r_d$, where $r_f$ is the format reward and $r_f=1$ if the response follows the think answer format, otherwise 0, $r_d$ is the concrete reward that $r_d=1$ if the predicted answer is equal to the ground truth number.

Agent VQA: $r=r_f+r_d+r_c$, where $r_f$ is the format reward and $r_f=1$ if the response follows the think answer format, otherwise 0, $r_d$ is the concrete reward that $r_d=1$ if the predicted action type is equal to the ground truth action type,\eg $\texttt{click, press\_home}$, $r_c$ is the continues reward for the predicted coordinates when the action type is click.
In this paper, we use the normalized coordinates ranging from 0-1 and set $\theta=0.14$ following~\cite{seeclick}.

Math VQA: $r=r_f+r_d$, where $r_f$ is the format reward and $r_f=1$ if the response follows the think answer format, otherwise 0, $r_d$ is the concrete reward that $r_d=1$ if the predicted answer is equal to the ground truth number.

\section{Limitations}
\label{sec: discussion}
Due to computational resources, our current work focuses on smaller-sized visual-language models like 3B and 7B, the proposed method has not been evaluated on even larger models (\eg 235B). 
We implement \ours on the open-domain VLMs; however, without access to the source code of proprietary VLMs like GPT-4o, the proposed method has not been implemented on them.

\section{Broader Impact} 
In this paper, we propose a simple yet effective method \ours, to cooperate SFT and RL stages by thought dropout. 
We improve the vanilla GRPO's performance by sampling minor code changes to teach the model to reason during the RL exploration stage selectively.
This enables a deeper understanding of RL in VLMs, inspiring flexible injection of prior knowledge into the SFT stage instead of manually creating rule-based rewards.
For social impact, this work has a certain impact on the RL research in the VLM and LLM.

\section{Dataset}
\label{app sec: dataset}
\noindent\textbf{General VQA.}  
The {CLEVR} dataset~\cite{johnson2017clevr} is designed to generate complex multi-step questions based on synthetic images, assessing a model’s true reasoning ability. 
It is a diagnostic dataset that includes 100,000 rendered images and approximately one million automatically generated questions, of which 853,000 are unique. 
The dataset features challenging questions involving counting, comparison, logical reasoning, and memory storage, while the images depict simple 3D shapes.
In contrast to the original CLEVR dataset, Super-CLEVR~\cite{li2023super} introduces more complex visual components and offers better control over the factors contributing to domain shift. 
For our experiments, we select a subset of 1,000 datasets that contain only counting problems for training. 
We evaluate the model's performance on test sets by selecting 200 samples from CLEVR that were not seen in the training set, as well as 200 counting problems from the out-of-distribution Super-CLEVR dataset.

\noindent\textbf{Math VQA.} {GeoQA}~\cite{chen2021geoqa} is a large-scale geometric question answering dataset that contains 4,998 geometric problems collected from real math exams in Chinese middle school.
Each problem is accompanied by annotated programs illustrating the solution process. While this dataset features multiple-choice questions, we increase the difficulty in this paper by removing the answer choices and requiring the model to generate the answers directly. We select a subset of 1k problems that involve computing angles and side lengths for training and test the model on this training set.

\noindent\textbf{GUI Agent.}  
{AITZ}~\cite{aitz} is a dataset designed for the graph user interface (GUI) navigation task derived from the large-scale mobile benchmark Android-in-the-wild (AITW~\cite{aitw}).
It features a unique annotation called chain-of-action thought (CoAT),
establishing a connection between perception—specifically, the understanding of screen layouts and UI elements—and cognition, which involves action decision-making. 
The AITZ dataset includes 2,504 operational trajectories that encompass 18.6K real-world intentions. 
Additionally, it is categorized into five subsets based on application domains: General, Install, GoogleApps, Single, and WebShopping. 
We train the model using the General domain with a dataset of randomly selected 1k examples and evaluate its performance on the corresponding test sets, as well as on other out-of-distribution domains.

\section{Setup}
\label{app sec: setup}
We use Llamafactory~\cite{zheng2024llamafactory} for the SFT stage with full parameters, and the training time is no longer than 15 minutes for both Qwen2.5-VL-3B/7B models.
We set $\theta=0.14$ following~\cite{seeclick}.
We use vLLM~\cite{vllm} and the zero1\_no\_optimizer GRPO settings to optimize further:

\begin{table*}[h]
    \centering
    \caption{Training Parameters for the first SFT of \ours }
    \begin{tabular}{|l|l|}
        \hline
        \textbf{Parameter} & \textbf{Value} \\ \hline
        cutoff\_len &  2048 \\ \hline
        per\_device\_train\_batch\_size & 8 \\ \hline
        gradient\_accumulation\_steps & 1 \\ \hline
        learning\_rate & 1.0e-5 \\ \hline
        lr\_scheduler\_type & cosine \\ \hline
        warmup\_ratio & 0.1 \\ \hline
        epoch & 2 \\ \hline
    \end{tabular}
    \vspace{-5mm}
    \label{tab:training_parameters}
\end{table*}

\begin{table*}[h]
    \centering
    \caption{Training Parameters for the second GRPO stage of \ours in general/agent }
    \begin{tabular}{|l|l|}
        \hline
        \textbf{Parameter} & \textbf{Value} \\ \hline
        max\_prompt\_length & 4096 \\ \hline
        max\_completion\_length & 2048 \\ \hline
        per\_device\_train\_batch\_size & 1 \\ \hline
        gradient\_accumulation\_steps & 2 \\ \hline
        learning\_rate & 1e-6 \\ \hline
        lr\_scheduler\_type & constant \\ \hline
        bf16 & true \\ \hline
        $\beta$ & 0.04 \\ \hline
        gradient\_checkpointing & true \\ \hline
        attn\_implementation & flash\_attention\_2 \\ \hline
        min\_pixels & 3136 \\ \hline
        max\_pixels & 501760 \\ \hline
        temperature & 1.0 \\ \hline
        num\_generations & 8\\ \hline
        step & 100 \\ \hline
    \end{tabular}
    
    \label{tab:training_parameters}
    \vspace{-5mm}
\end{table*}

\begin{table*}[h]
    \centering
    \caption{Training Parameters for the second GRPO stage of \ours in math }
    \begin{tabular}{|l|l|}
        \hline
        \textbf{Parameter} & \textbf{Value} \\ \hline
        max\_prompt\_length & 4096 \\ \hline
        max\_completion\_length & 2048 \\ \hline
        per\_device\_train\_batch\_size & 1 \\ \hline
        gradient\_accumulation\_steps & 2 \\ \hline
        learning\_rate & 1e-6 \\ \hline
        lr\_scheduler\_type & constant \\ \hline
        bf16 & true \\ \hline
        $\beta$ & 0.04 \\ \hline
        attn\_implementation & flash\_attention\_2 \\ \hline
        min\_pixels & 3136 \\ \hline
        max\_pixels & 501760 \\ \hline
        temperature & 1.0 \\ \hline
        num\_generations & 4\\ \hline
        step & 300 \\ \hline
    \end{tabular}
    
    \label{tab:training_parameters}
\end{table*}

\section{Experiments}

\subsection{\ours on Math--GeoQA}
\label{app: ton on math}

\begin{figure}[h!]
    \centering
    \includegraphics[width=\linewidth]{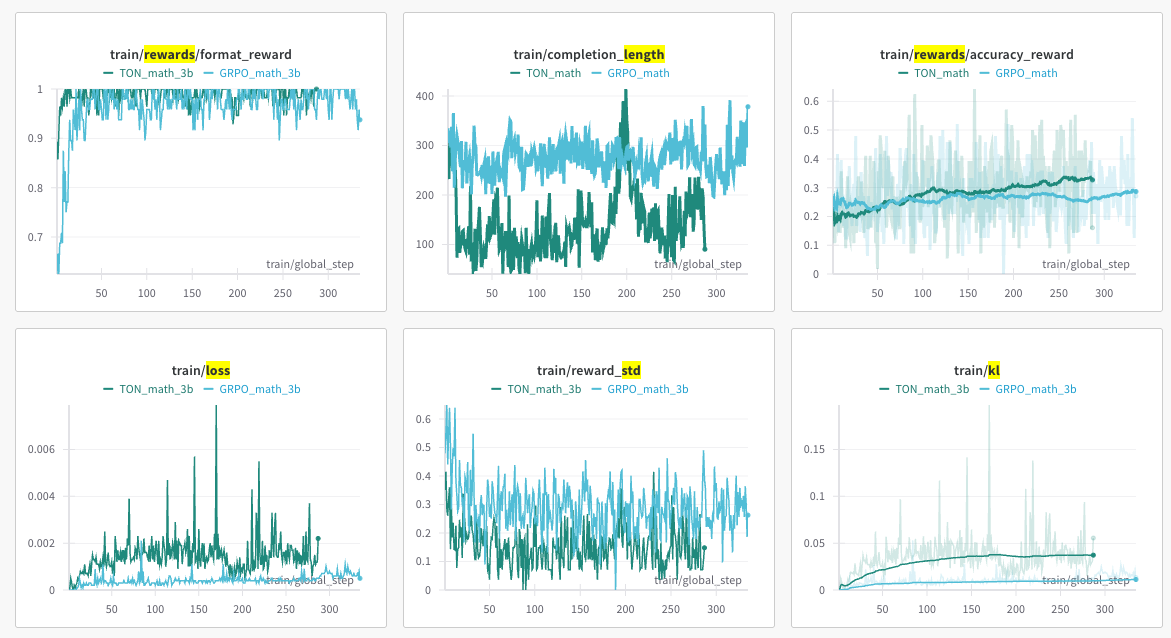}
    \caption{\ours and GRPO visualization during the training process on Qwen2.5-VL-3B on GeoQA.}
    \label{app fig: math 3b}
\end{figure}

\begin{figure}[h!]
    \centering
    \includegraphics[width=\linewidth]{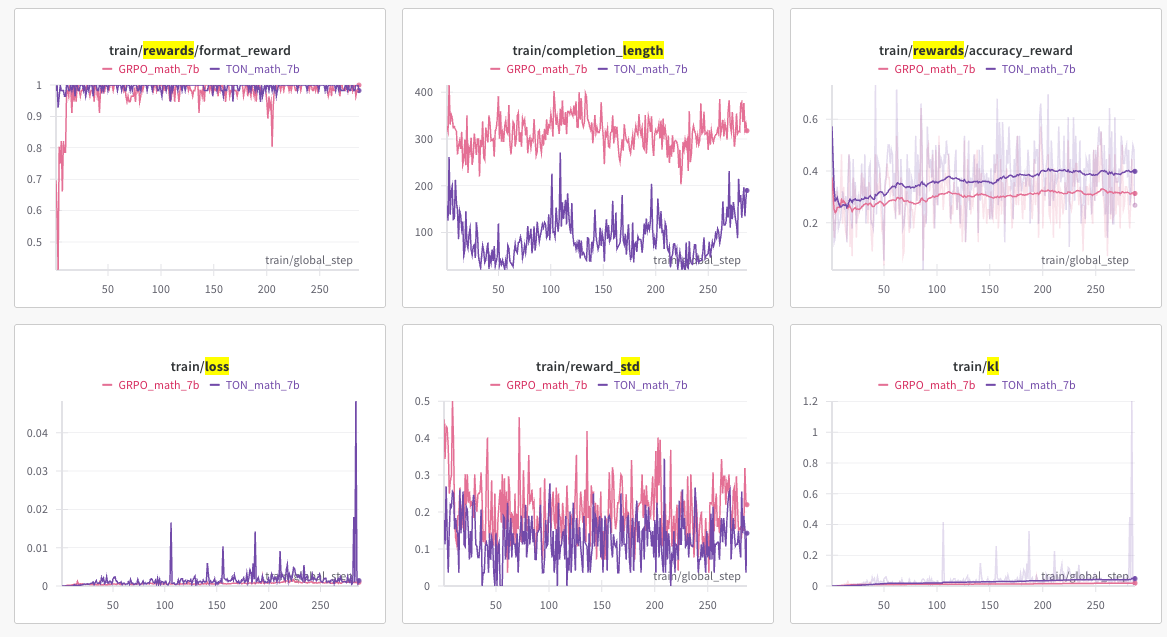}
    \caption{\ours and GRPO visualization during the training process on Qwen2.5-VL-7B on GeoQA.}
    \label{app fig: math 7b}
\end{figure}

Figure~\ref{app fig: math 3b} \& \ref{app fig: math 7b} illustrate the progression of various variables throughout the training process.

\newpage
\subsection{\ours on Counting--CLEVR}
\label{app: ton on general}

\begin{figure}[h!]
    \centering
    \includegraphics[width=\linewidth]{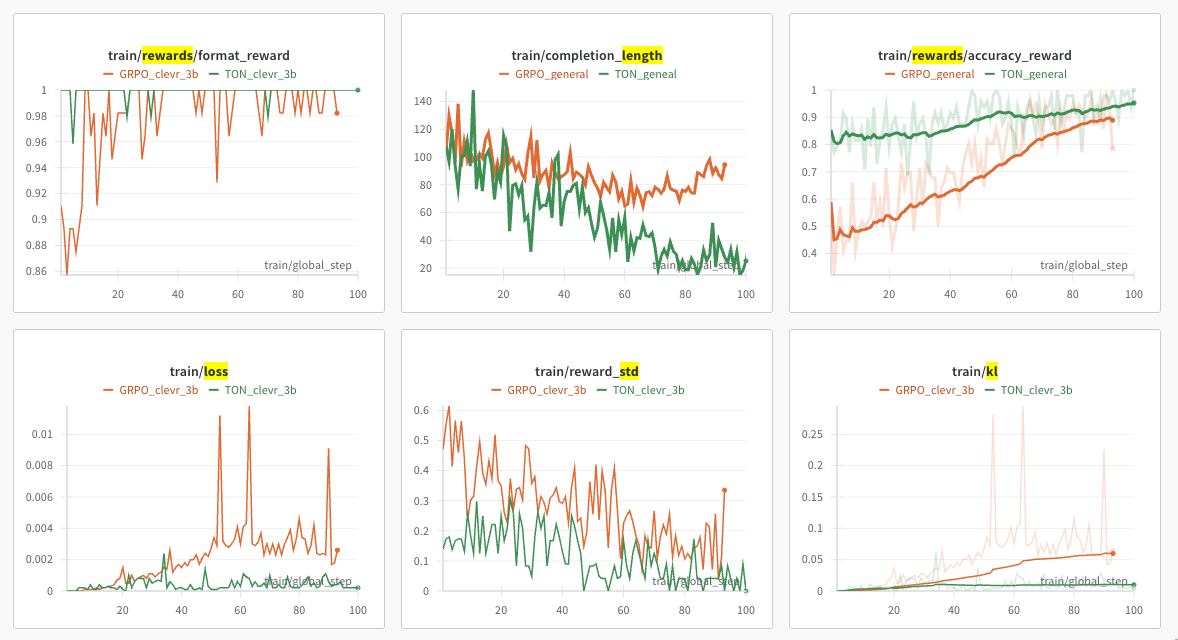}
    \caption{\ours and GRPO visualization during the training process on Qwen2.5-VL-3B on CLEVR.}
    \label{app fig: count}
\end{figure}

Figure~\ref{app fig: count} illustrates the progression of various variables throughout the training process.

\subsection{\ours on Mobile Agent--AITZ}
\label{app: ton on agent}

\begin{figure}[h!]
    \centering
    \includegraphics[width=\linewidth]{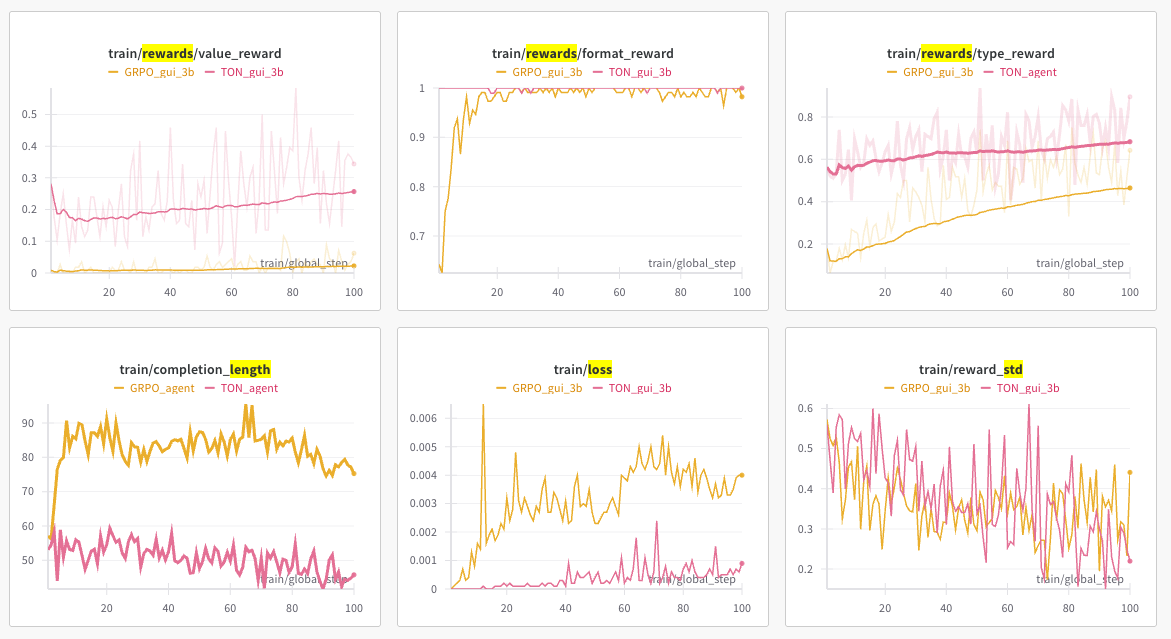}
    \caption{\ours and GRPO visualization during the training process on Qwen2.5-VL-3B on AITZ.}
    \label{app fig: gui 3b}
\end{figure}

\begin{figure}[h!]
    \centering
    \includegraphics[width=\linewidth]{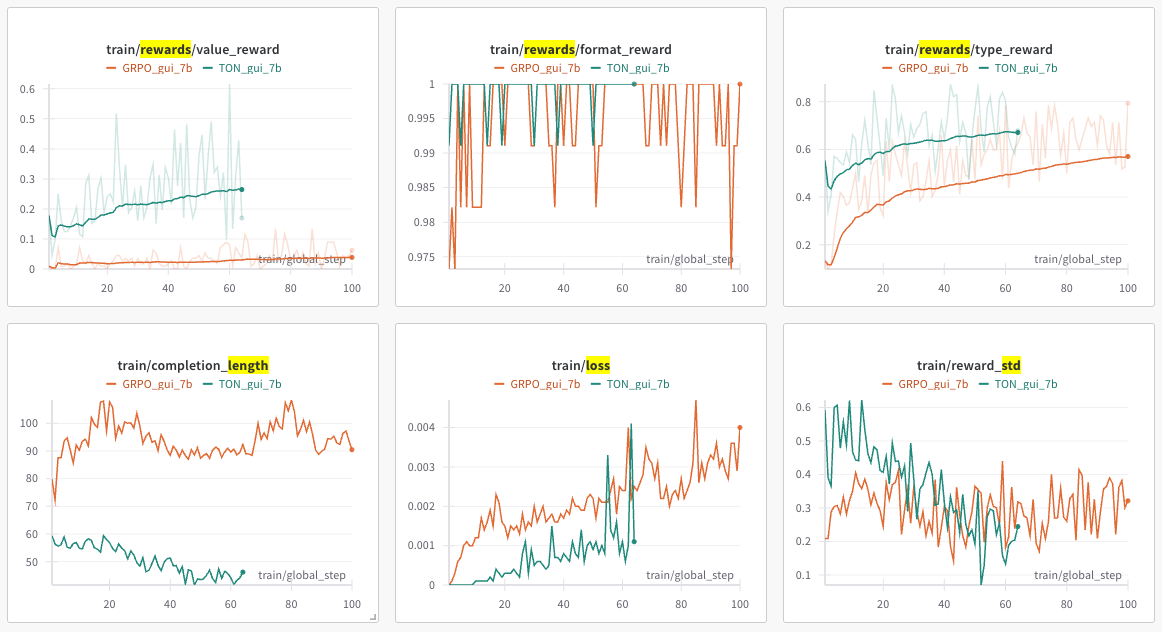}
    \caption{\ours and GRPO visualization during the training process on Qwen2.5-VL-7B on AITZ.}
    \label{app fig: gui 7b}
\end{figure}

Figure~\ref{app fig: gui 3b} \& \ref{app fig: gui 7b} illustrate the progression of various variables throughout the training process.

\newpage
\subsection{OOD Performance of \ours on CLEVR}
\label{app sub: ood performance on general}
Table~\ref{app tab: ood general} compares the IID and OOD performance of \ours and vanilla GRPO. \ours demonstrates superior performance in both IID and, particularly, OOD scenarios under easy reasoning tasks, helping to avoid overfitting to the training set of vanilla GRPO.

\begin{table*}[h!]
    \centering
  \caption{Qwen2.5-VL-3B on the IID domain CLEVR and OOD domain Super-CLEVR. 
  }
    \resizebox{0.5\linewidth}{!}{
  \begin{tabular}{cccc}
    \toprule
     & & CLEVR  & Super-CLEVR \\
     &  think & acc & acc \\
    \midrule
    base & & 64 & 57.3 \\
    SFT & \cmark & 88.5 & 13.17 \\
    \midrule
    GRPO & \cmark &  93.5 & 51.9\\
    \ours & ours & 98.5 & 62.79\\
    \bottomrule
  \end{tabular}
  }
  \label{app tab: ood general}
\end{table*}


\newpage
\subsection{Different Thought Dropout Probabilities}
\label{app sec: different dropout ratio}
\begin{figure}[h!]
    \centering
    \includegraphics[width=\linewidth]{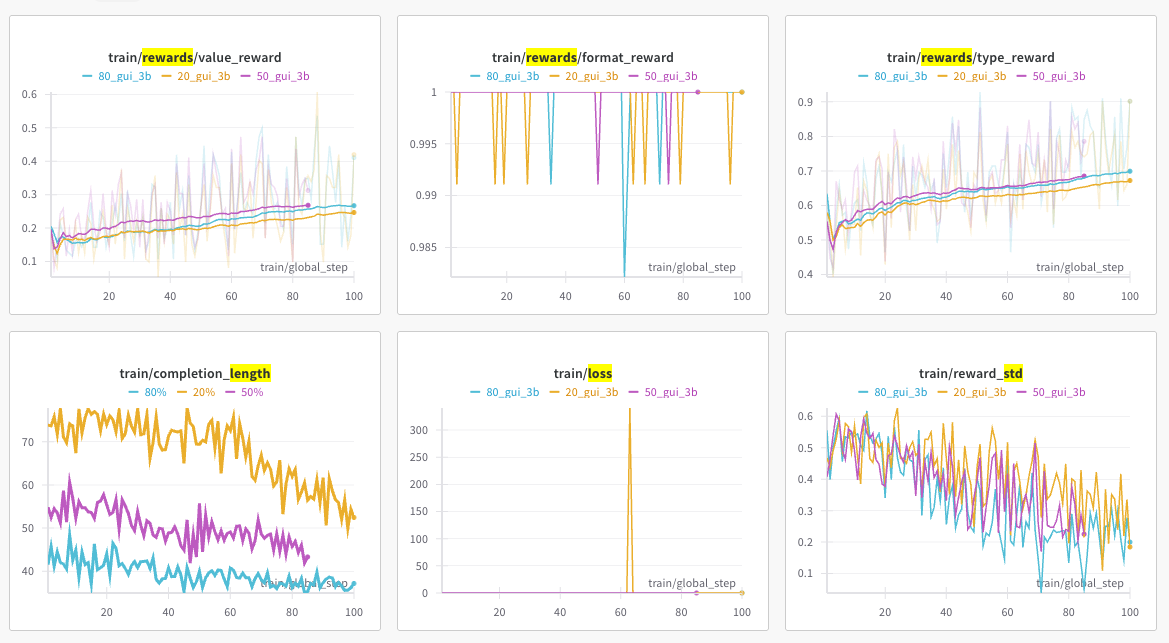}
    \caption{GRPO visualization during the training process on Qwen2.5-VL-3B on AITZ under dropout probabilities 20\%, 50\%, 80\%.}
    \label{app fig: skip train}
\end{figure}

Figure~\ref{app fig: skip train} illustrates the progression of various variables throughout the training process under different dropout probabilities.

\subsection{Skip-thought Ratio on Different benchmarks}
\label{app sec: skip thought ratio}

\begin{figure}[h!]
\centering
\begin{subfigure}[b]{0.32\textwidth}
    \includegraphics[width=\linewidth]{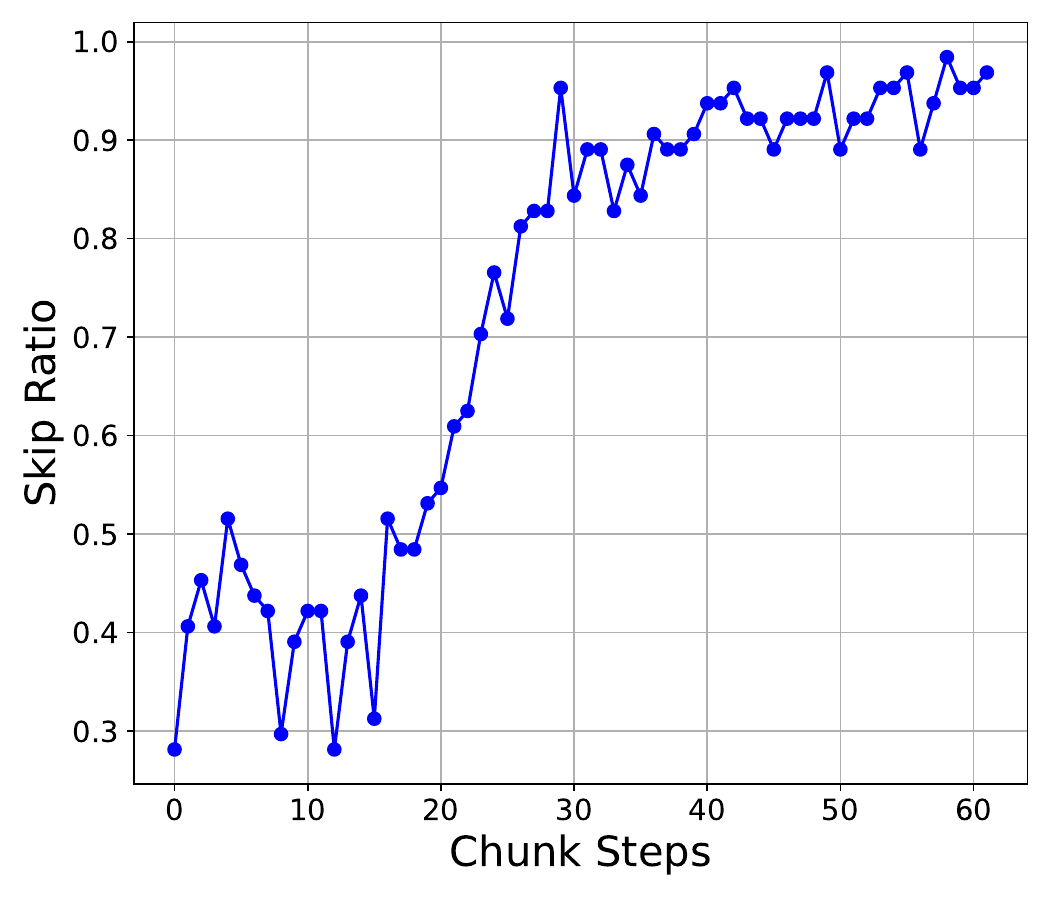}
    \caption{
    Counting-CLEVR}
    \label{app fig: skip general}
\end{subfigure}
\hfill
\begin{subfigure}[b]{0.32\textwidth}
    \includegraphics[width=\linewidth]{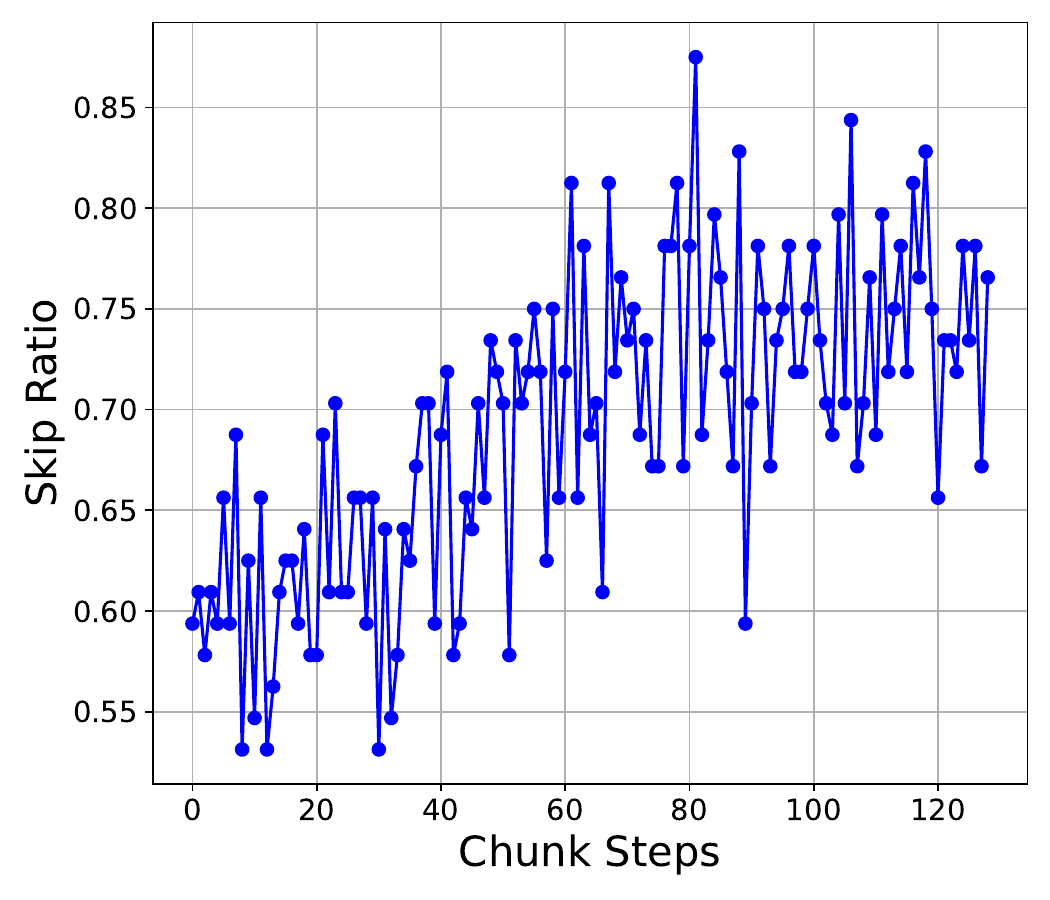}
    \caption{AITZ}
    \label{app fig: skip agent}
\end{subfigure}
\hfill
\begin{subfigure}[b]{0.32\textwidth}
    \includegraphics[width=\linewidth]{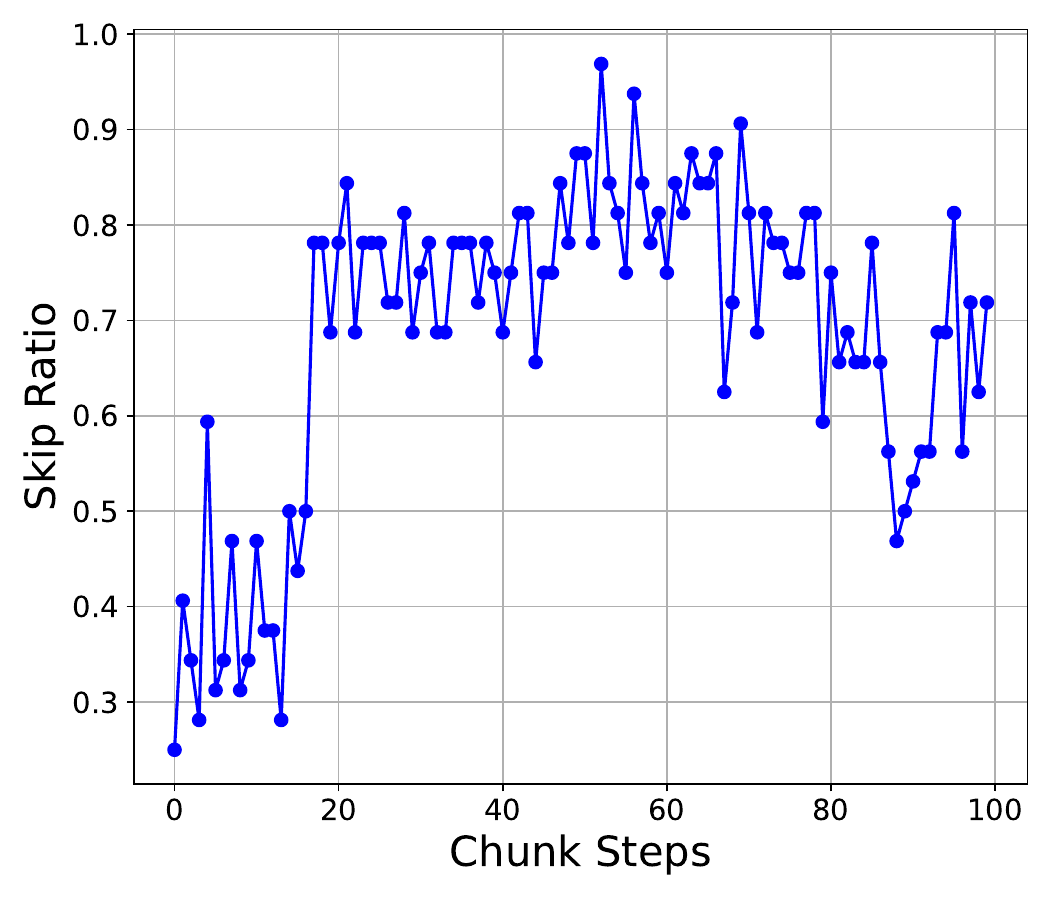}
    \caption{GeoQA}
    \label{app fig: skip math}
\end{subfigure}
\caption{\textbf{Skip Ratio of the output thinking} during our \ours training on three benchmarks. 
}
\label{app fig:  skip ratio}
\end{figure}

Figure~\ref{app fig:  skip ratio} illustrates the skip-thought ratios under \ours throughout the training process under different VQA benchmarks.


\newpage
\subsection{Prompt v.s. SFT on different benchmarks}
\label{app sec: prompt sft}
\begin{figure}[h]
    \centering
    \includegraphics[width=\linewidth]{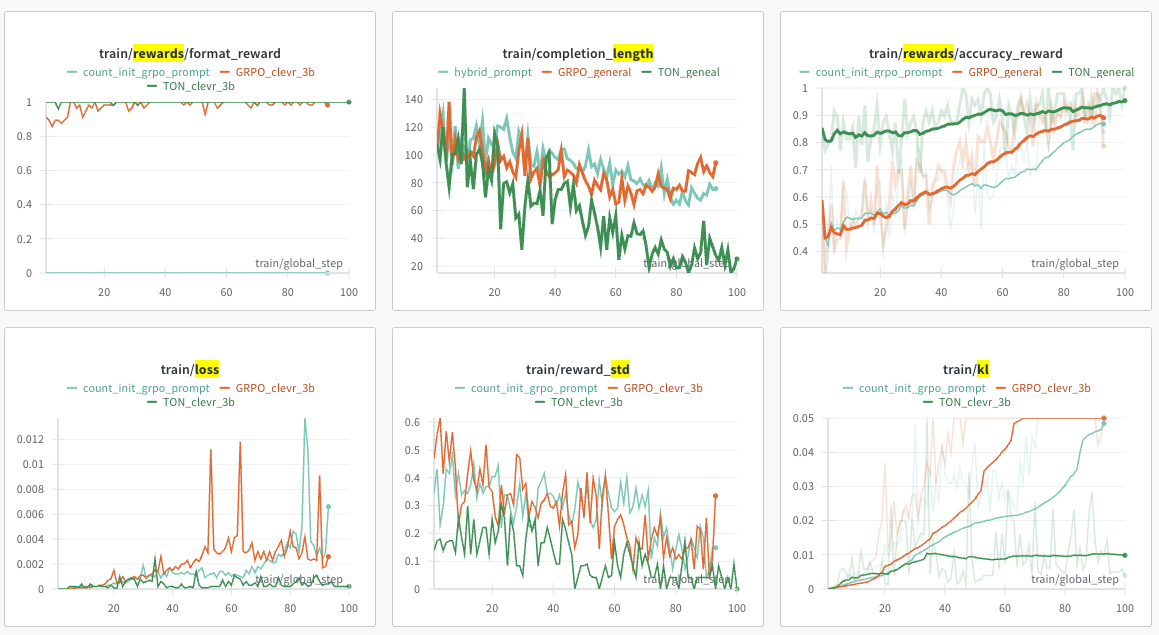}
    \caption{hybrid prompt v.s. SFT visualization during the training process on Qwen2.5-VL-3B on clevr.}
    \label{app fig: prompt clevr}
\end{figure}

\begin{figure}
    \centering
    \includegraphics[width=\linewidth]{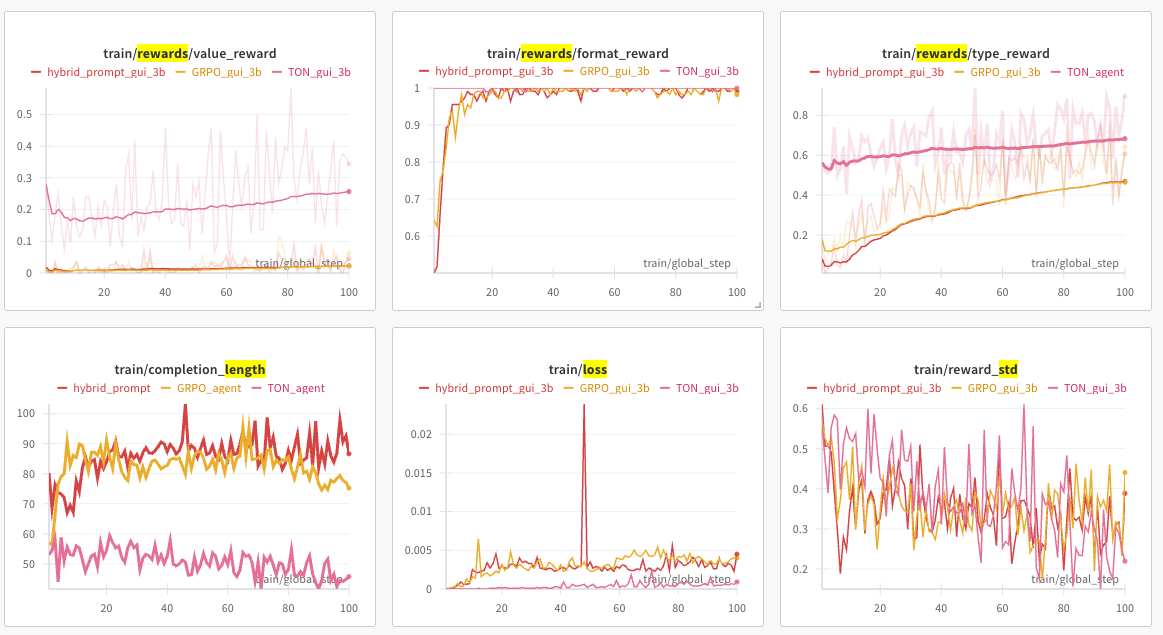}
    \caption{hybrid prompt v.s. SFT visualization during the training process on Qwen2.5-VL-3B on AITZ.}
    \label{app fig: prompt gui}
\end{figure}

\begin{figure}
    \centering
    \includegraphics[width=\linewidth]{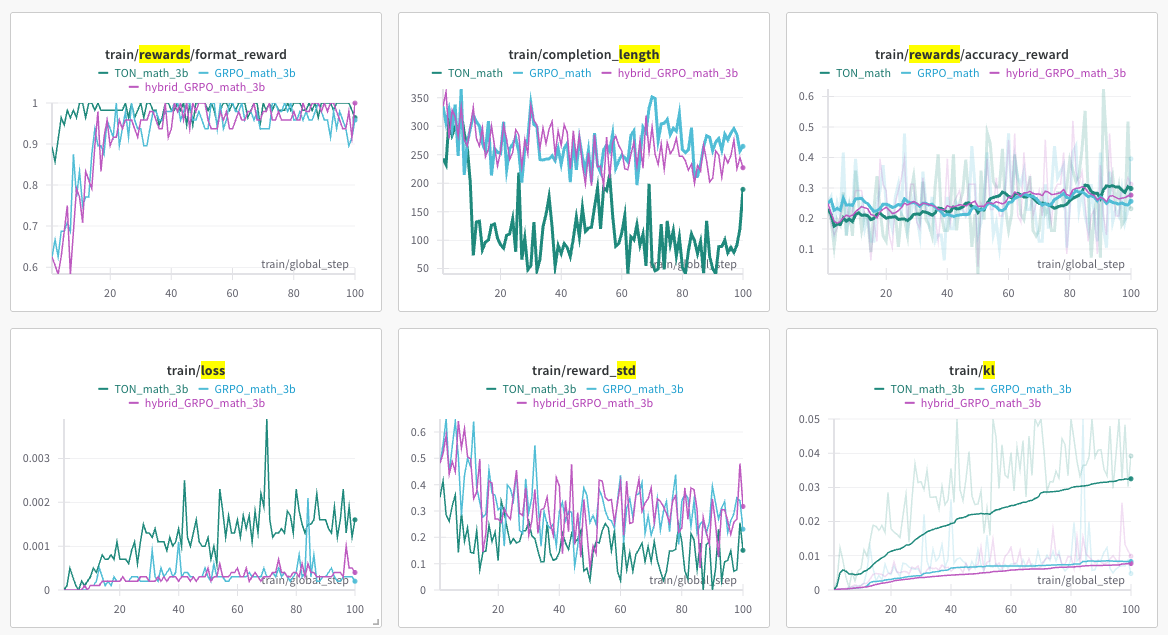}
    \caption{hybrid prompt v.s. SFT visualization during the training process on Qwen2.5-VL-3B on GeoQA.}
    \label{app fig: prompt math}
\end{figure}

Figure~\ref{app fig: prompt clevr} \& ~\ref{app fig: prompt gui} \& ~\ref{app fig: prompt math} illustrate the progression of various variables throughout the training process between injecting the skip-thought during the prompt and the SFT stage.

\newpage

\subsection{Visualization Examples}
\label{app sec: examples}

\begin{figure}[h!]
    \centering
    \includegraphics[width=\linewidth]{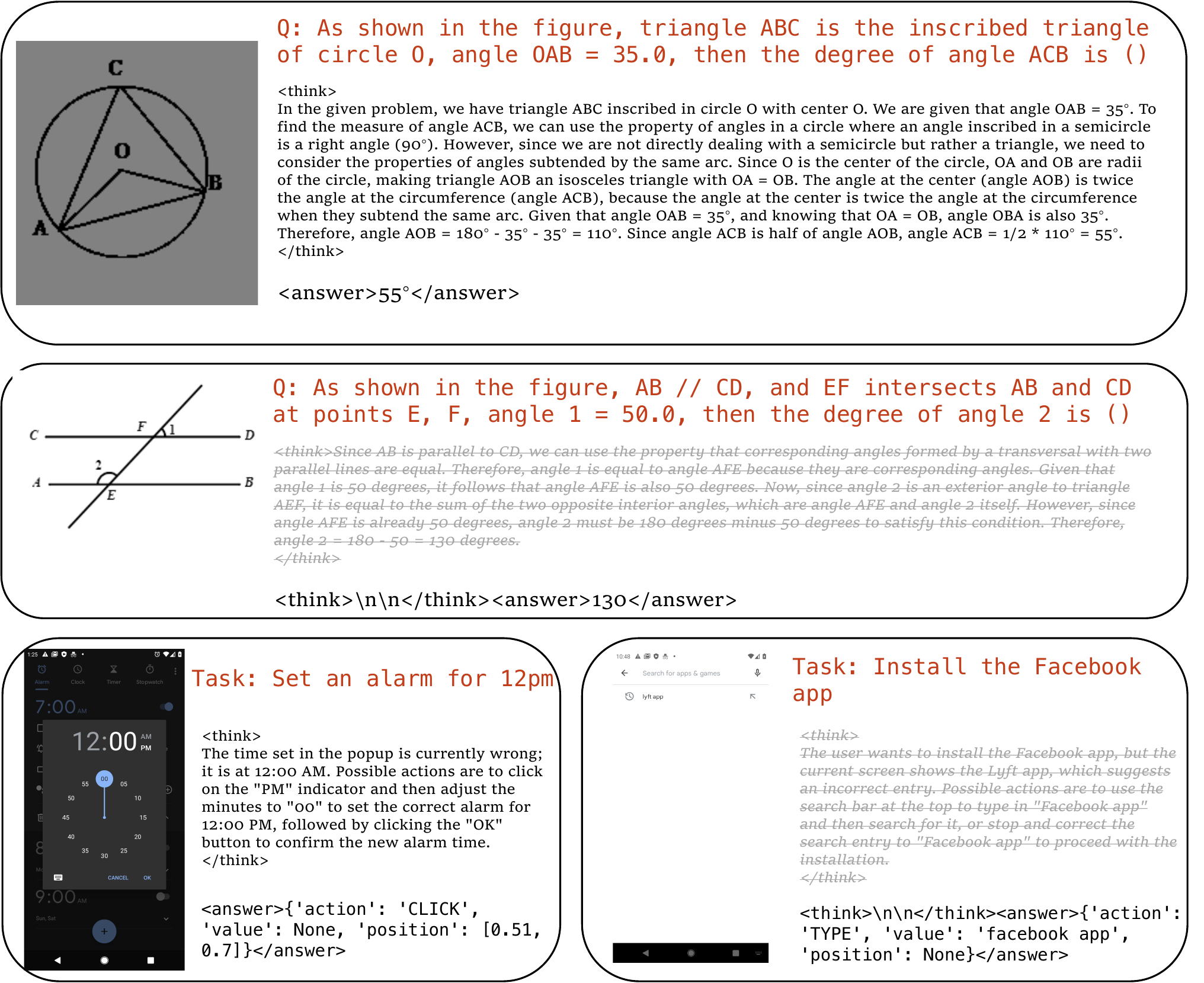}
    \caption{Examples of \ours on math VQA and GUI agent VQA settings.}
    \label{fig:enter-label}
\end{figure}

\begin{table}[h!]
  \begin{minipage}{0.99\textwidth}
\centering
\vspace{-4mm}
\scalebox{0.88}{
\begin{tabular}{l p{5.4cm} p{6.0cm} }
\toprule
 \multicolumn{3}{l}{\bf Counting examples from CLEVR~\cite{johnson2017clevr}}  \\
\midrule
&  \includegraphics[height=3.5cm]{figures/example/count_easy.png} 
&  \includegraphics[height=3.5cm]{figures/example/count_hard.png} \\
\midrule
Question & How many items are there in the image? & How many items are there in the image?
\\ \midrule
GRPO's outputs & 
<think>The image shows three objects: a large green cylinder from 1919, a large teal sphere, and a large golden sphere. Both the teal and golden spheres appear to be in a translucent state, indicating that they are metallic. The green cylinder is plain and opaque, without any reflective properties.
1. Green Cylinder: Transparent (metallic)
2. Teal Sphere: Transparent (metallic)
3. Golden Sphere: Transparent (metallic)
The question seems to ask for the count of items, and considering that all three objects are distinguishable from each other as separate items in the image, the count is three.
</think>

<answer>3</answer>

& 
<think>
- I count six distinct objects in the picture: a large green matte cylinder, a blue matte cylinder, a large green reflective sphere, a large blue matte cube, a small blue matte cube, a light blue rubber object, a shiny silver cube, a large reflective blue object, a silver or dark green reflective sphere, a shiny purple metallic cylinder, and a blue matte sphere.
- I observe the reflective surface properties of the objects around the central large green sphere.
- I also note the size and color variety of each object as described by the visual cues.
</think>

<answer>10</answer>
\\ \midrule
TON's outputs & <think>\textbackslash n\textbackslash n</think>

<answer>3</answer> & 
<think>Okay, let's check this out. The user provided a image with several objects and their properties. Each entry starts with "A" followed by attributes like size, color, material, shape, rotation, 3D coordinates, and pixel coordinates.

First, I need to count how many items there are. Let me list each entry one by one.

1. "A small purple rubber sphere..." That's the first item.
2. "A large yellow rubber cube..." Second item.
\textcolor{blue}{3. "A small gray rubber cube..." Third.}
4. "A large green rubber cylinder..." Fourth.
5. "A small blue metal cube..." Fifth.
6. "A small green metal sphere..." Sixth.
7. "A large brown metal cube..." Seventh.
8. "A small green rubber cylinder..." Eighth.

Wait, that's eight distinct entries. Each one has unique combinations of attributes. I can't see any duplicates. Each has different values for color, size, material, rotation, or coordinates, so they're separate. So yes, there are eight items in total.
</think>

<answer> 8 </answer>
\\ \bottomrule
\end{tabular}
}
\vspace{1mm}
\captionof{table}{
Counting example from CLEVR~\cite{johnson2017clevr}. 
Although the question is simple, the two conditioned images differ significantly in difficulty: the left image is clearly easier, while the right involves greater complexity due to object occlusion. TON demonstrates selective activation of reasoning—engaging thought only when needed—whereas GRPO generates reasoning traces for both cases indiscriminately.}

\label{tab:clevr example}  
  \end{minipage}
\end{table}

\subsection{Prompt for AITZ task}

\begin{AIbox}{AITZ System Prompt}

You are an assistant trained to navigate the mobile phone. 
Given a task instruction, a screen observation, and an action history sequence, output the next action and wait for the next observation. \\
Here is the action space:
\begin{itemize}
    \item 1. `CLICK`: Click on an element, value is not applicable and the position [x,y] is required. 
    \item 2. `TYPE`: Type a string into an element, value is a string to type and the position is not applicable.
    \item 3. `SCROLL UP`: Scroll up for the screen.
    \item 4. `SCROLL DOWN`: Scroll down for the screen.
    \item 5. `SCROLL LEFT`: Scroll left for the screen.
    \item 6. `SCROLL RIGHT`: Scroll right for the screen.
    \item 7. `PRESS BACK`: Press for returning to the previous step, value and position are not applicable.
    \item 8. `PRESS HOME`: Press for returning to the home screen, value and position are not applicable.
    \item 9. `PRESS ENTER`: Press for submitting the input content, value and position are not applicable.
    \item 10. `STATUS TASK COMPLETE`: Indicate the task is completed, value and position are not applicable.
\end{itemize}
Format the action as a dictionary with the following keys:
\{'action': 'ACTION TYPE', 'value': 'element', 'position': [x,y]\}\\
If value or position is not applicable, set it as `None`.
Position represents the relative coordinates on the screenshot and should be scaled to a range of 0-1.
\end{AIbox}

\begin{AIbox}{Think Prompt}
Please first thinks about the reasoning process in the mind and then provides the user with the action. The reasoning process and answer are enclosed within <think> </think> and <action> </action> tags, respectively, i.e.,
<think> reasoning process here </think><action> action here </action>
\end{AIbox}

\begin{AIbox}{Overall Prompt}
\{AITZ System Prompt\}\\
\{Think Prompt\}\\
Task: \{Task\} \\
<image> \\
Action history: \{Action history\} \\
**Next action:**    
\end{AIbox}

\newpage

\input{sec/supp}

%% file: sec/supp.tex

\section{Comprehensive Comparison of Length, Rewards, and Skip Ratio Across Three Benchmarks}

We present a comprehensive comparison of length, rewards, and skip ratio across three benchmarks. The results reveal a consistent trend: \ours reduces completion length and increases the skip ratio as rewards increase during training.

\begin{figure}[h!]
\centering
\begin{subfigure}[b]{0.32\textwidth}
    \includegraphics[width=\linewidth]{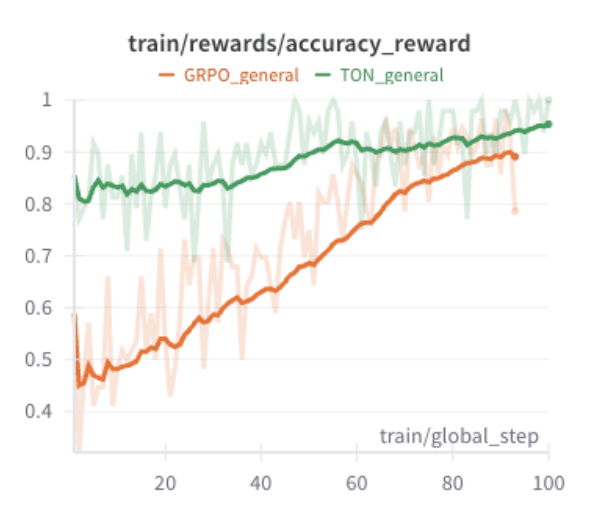}
    \caption{
    Counting-CLEVR}
    \label{app fig: skip general}
\end{subfigure}
\hfill
\begin{subfigure}[b]{0.32\textwidth}
    \includegraphics[width=\linewidth]{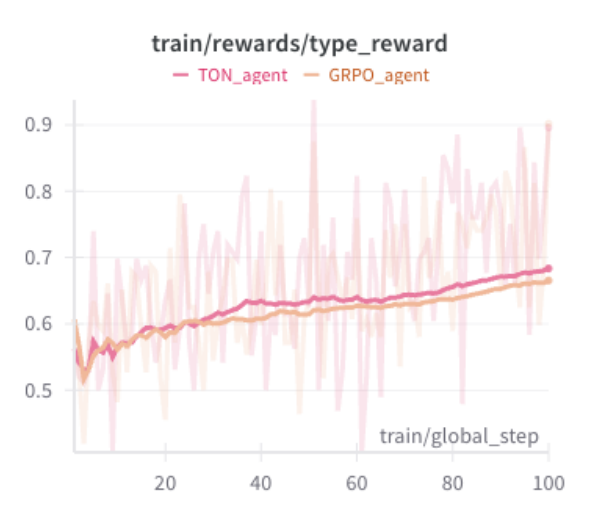}
    \caption{AITZ}
    \label{app fig: skip agent}
\end{subfigure}
\hfill
\begin{subfigure}[b]{0.32\textwidth}
    \includegraphics[width=\linewidth]{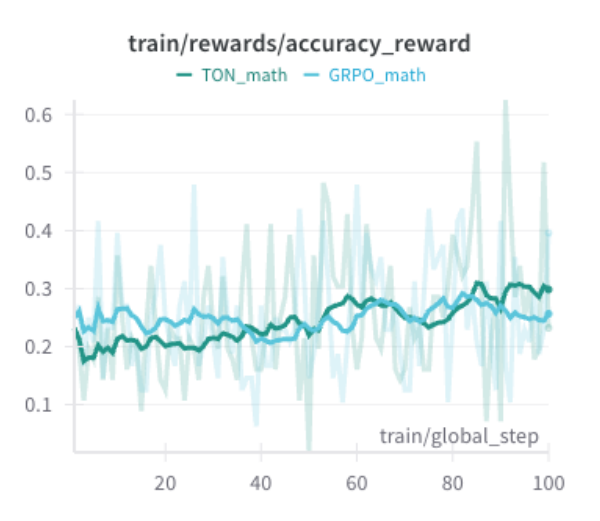}
    \caption{GeoQA}
    \label{app fig: skip math}
\end{subfigure}
\caption{\textbf{Rewards of the output} during our \ours training on three benchmarks. 
}
\label{app fig:  skip ratio}
\end{figure}

\begin{figure}[h!]
\centering
\begin{subfigure}[b]{0.32\textwidth}
    \includegraphics[width=\linewidth]{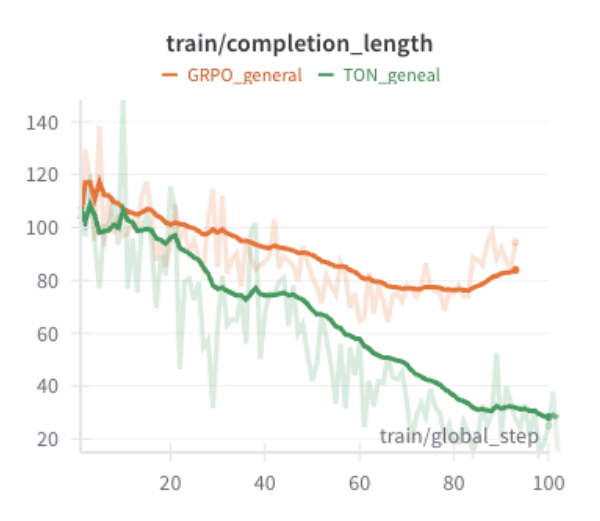}
    \caption{
    Counting-CLEVR}
    \label{app fig: skip general}
\end{subfigure}
\hfill
\begin{subfigure}[b]{0.32\textwidth}
    \includegraphics[width=\linewidth]{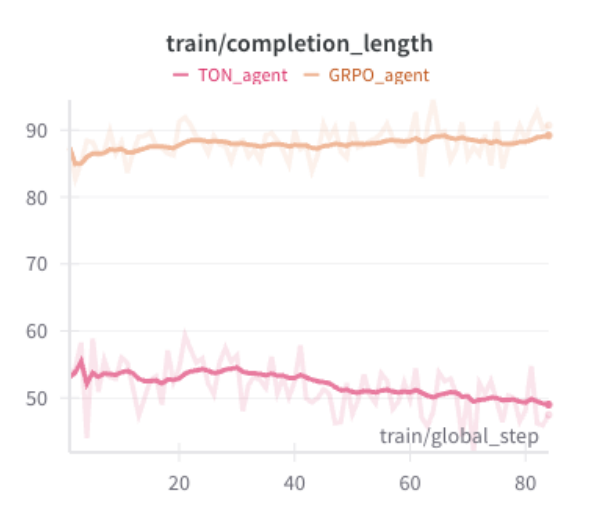}
    \caption{AITZ}
    \label{app fig: skip agent}
\end{subfigure}
\hfill
\begin{subfigure}[b]{0.32\textwidth}
    \includegraphics[width=\linewidth]{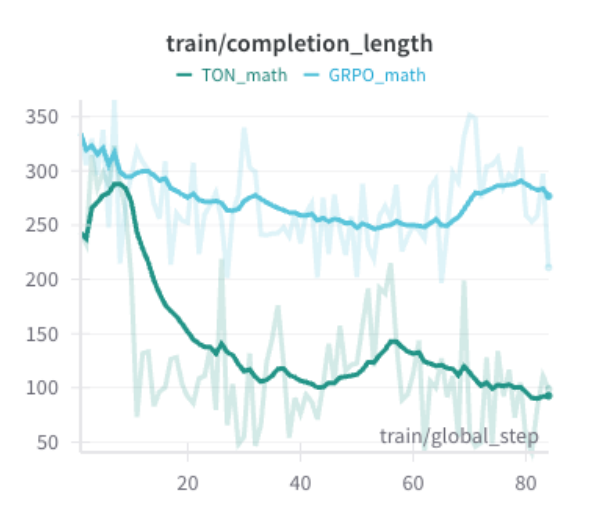}
    \caption{GeoQA}
    \label{app fig: skip math}
\end{subfigure}
\caption{\textbf{Completion length of the output} during our \ours training on three benchmarks. 
}
\label{app fig:  skip ratio}
\end{figure}

\begin{figure}[h!]
\centering
\begin{subfigure}[b]{0.32\textwidth}
    \includegraphics[width=\linewidth]{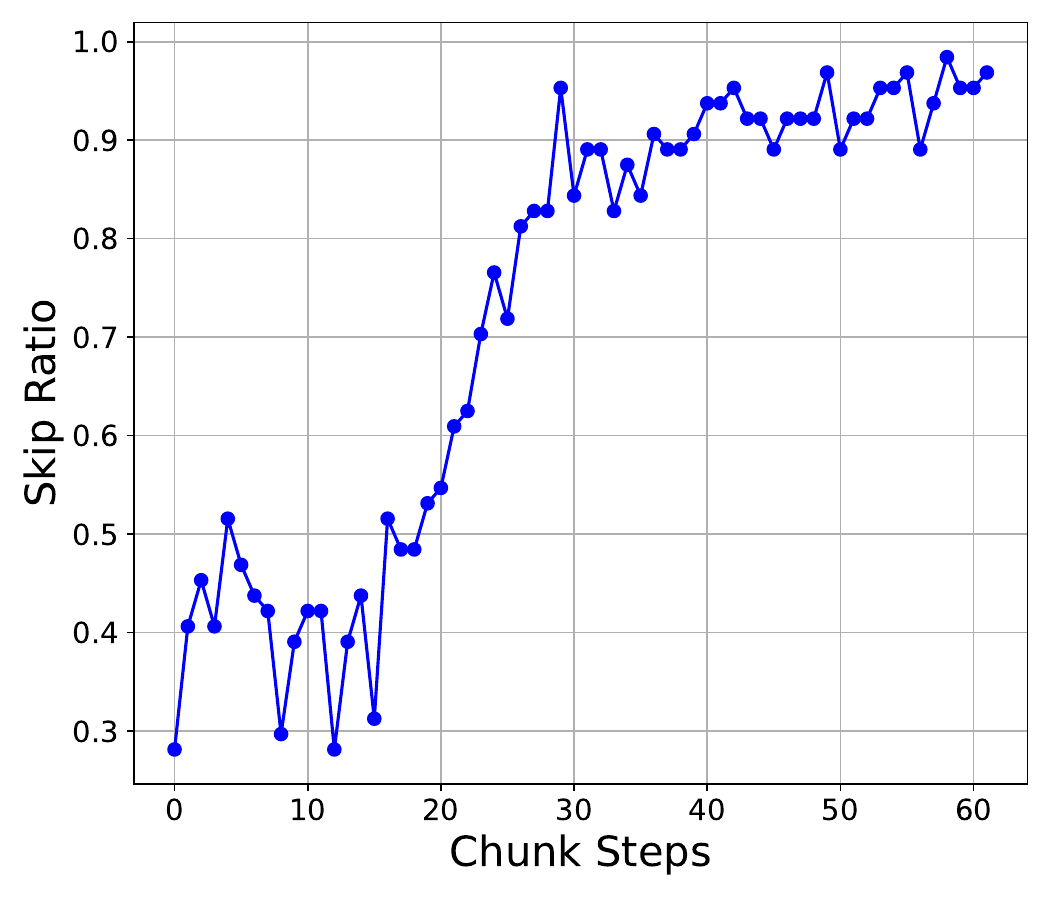}
    \caption{
    Counting-CLEVR}
    \label{app fig: skip general}
\end{subfigure}
\hfill
\begin{subfigure}[b]{0.32\textwidth}
    \includegraphics[width=\linewidth]{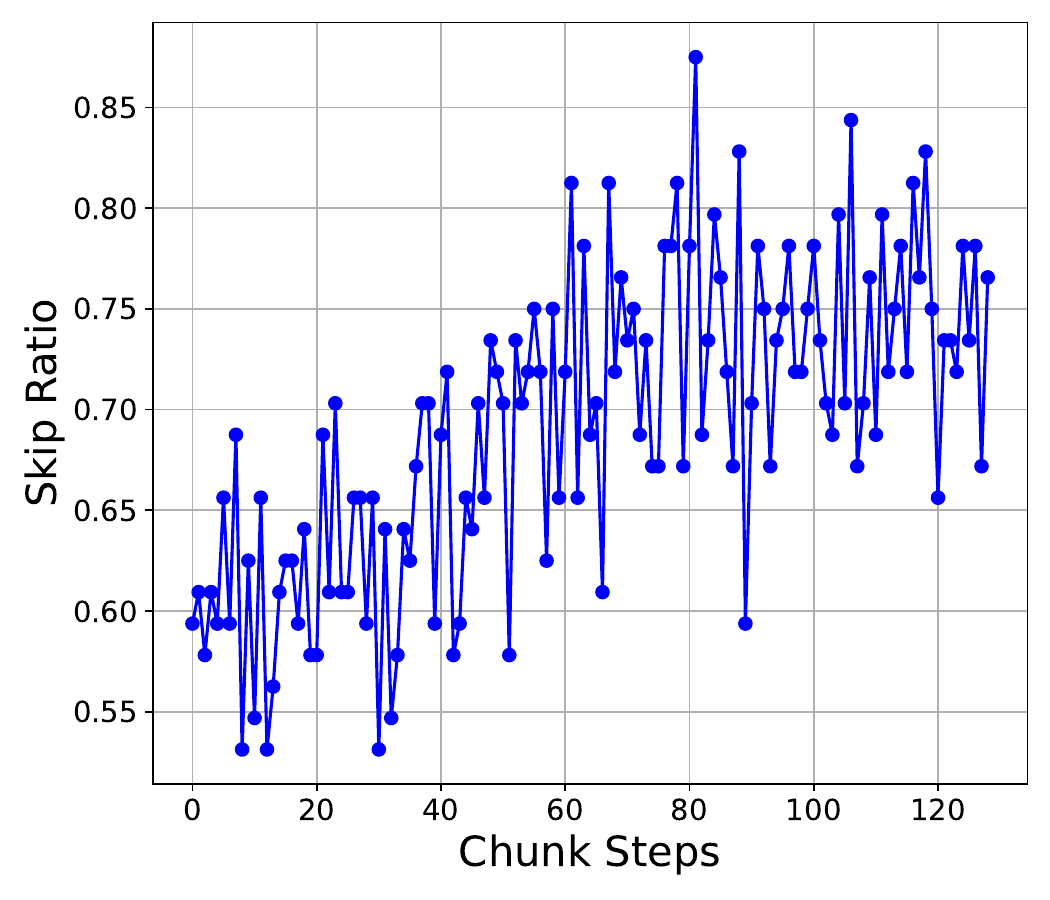}
    \caption{AITZ}
    \label{app fig: skip agent}
\end{subfigure}
\hfill
\begin{subfigure}[b]{0.32\textwidth}
    \includegraphics[width=\linewidth]{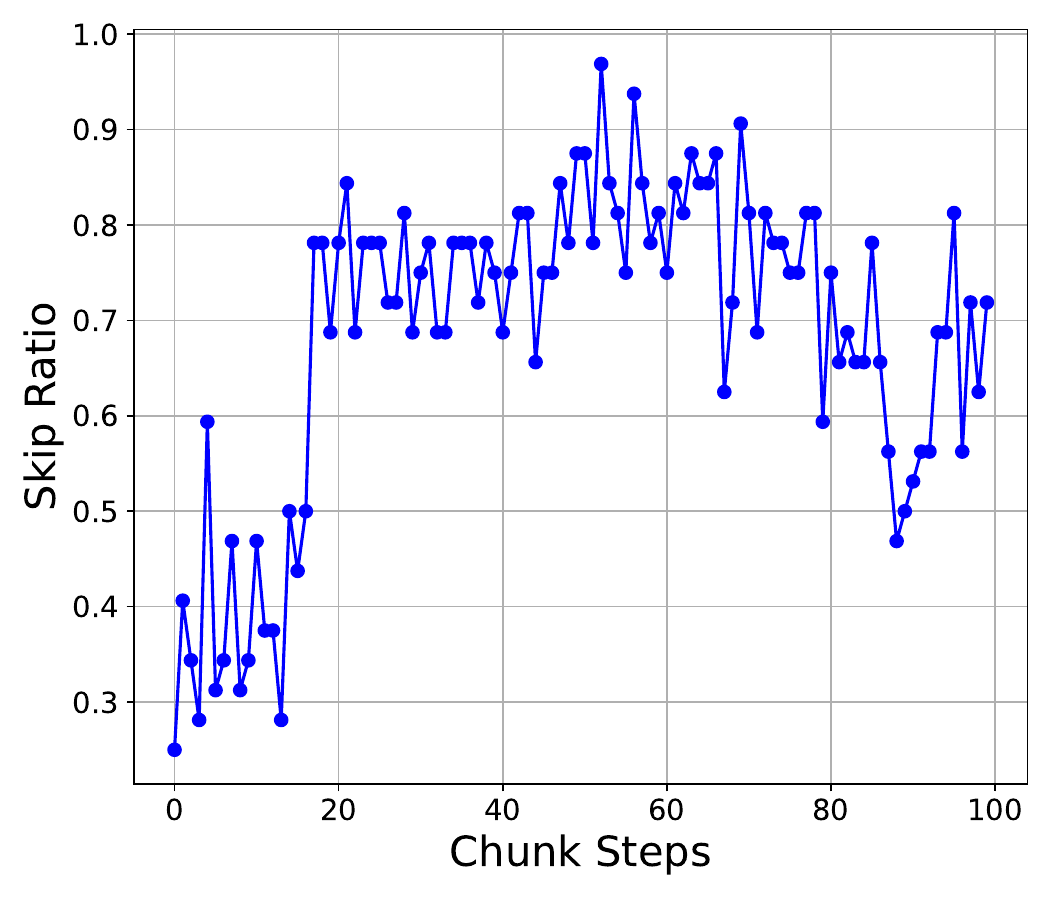}
    \caption{GeoQA}
    \label{app fig: skip math}
\end{subfigure}
\caption{\textbf{Skip ratio of the output thinking} during our \ours training on three benchmarks. 
}
\label{app fig:  skip ratio}
\end{figure}

\section{Reward for length rather than SFT}

We give the reward $r_l=1$ for the model if it outputs <think>\textbackslash n\textbackslash n<\textbackslash think>, and otherwise 0.
We observe that the length reward remains at 0 during the first 100 steps. The visualization of the entire training process is shown below, highlighting our proposed thought dropout in the SFT stage.

\begin{figure}[h!]
    \centering
    \includegraphics[width=\linewidth]{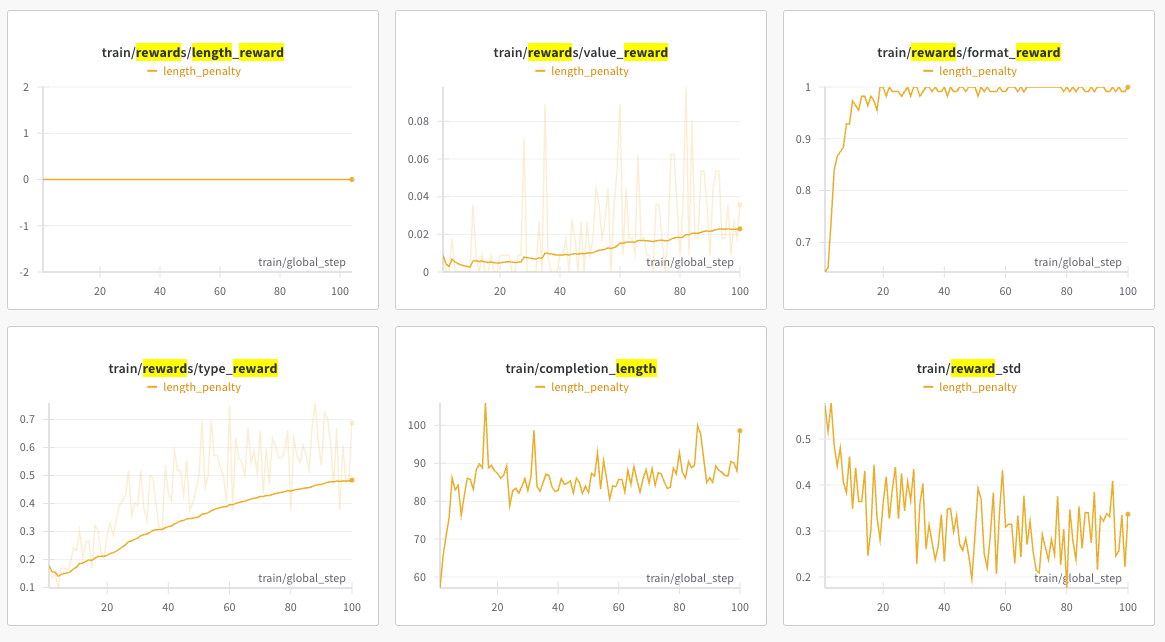}
    \caption{Length penalty rewards and completion length on AITZ datasets on Qwen2.5-VL-3B.}
    \label{fig:enter-label}
\end{figure}

\begin{figure}[h!]
    \centering
    \includegraphics[width=\linewidth]{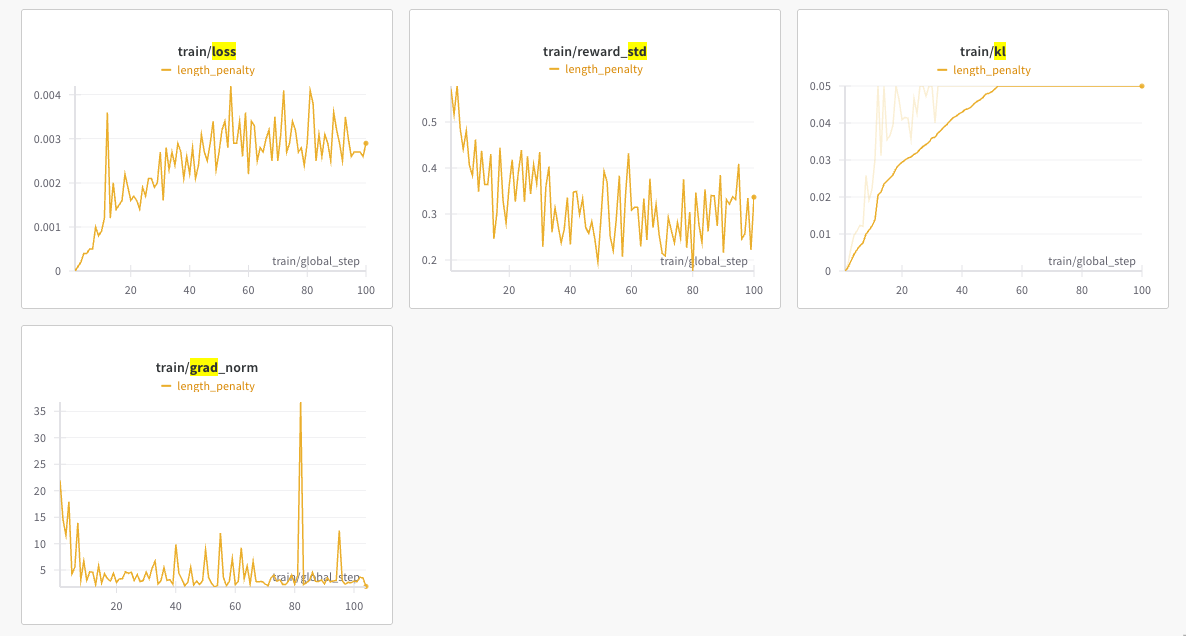}
    \caption{Length penalty rewards and completion length on AITZ datasets on Qwen2.5-VL-3B.}
    \label{fig:enter-label}
\end{figure}


\section{More cases}
We give more cases to show the effectiveness and efficiency of our proposed \ours.
\begin{figure}
    \centering
    \includegraphics[width=\linewidth]{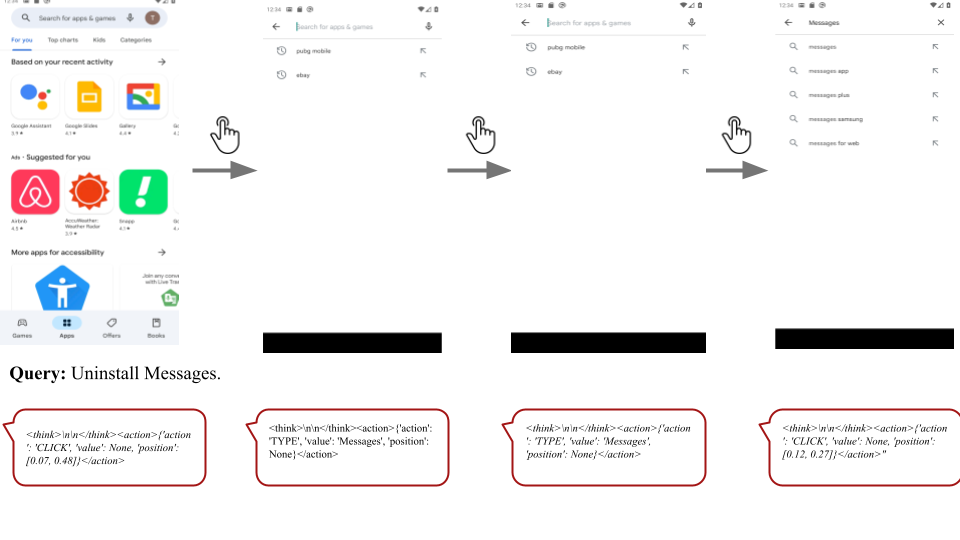}
    \caption{\ours applied to the AITZ out-of-distribution domain: the task is to uninstall the messaging apps. \ours performs well without extensive reasoning.}
    \label{fig:enter-label}
\end{figure}

\begin{figure}
    \centering
    \includegraphics[width=\linewidth]{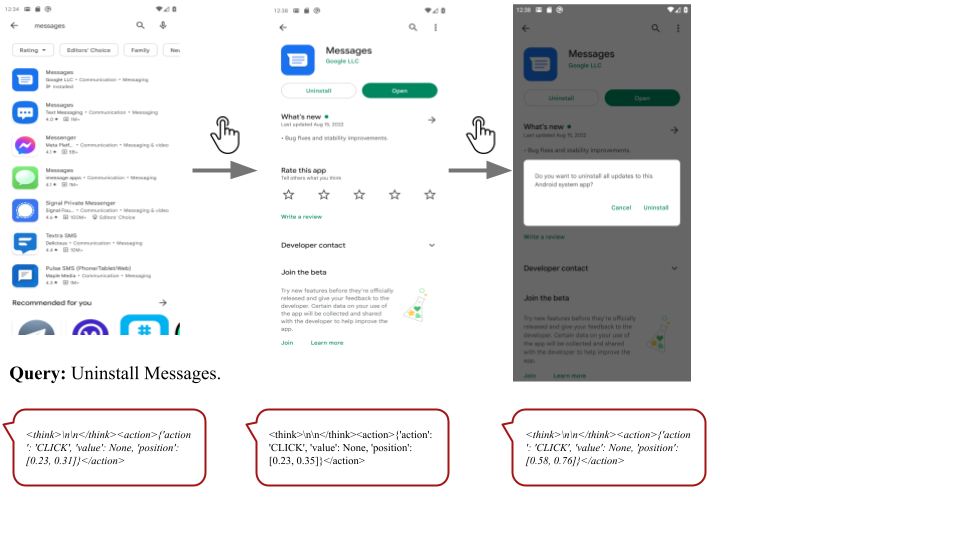}
    \caption{\ours applied to the AITZ out-of-distribution domain: the task is to uninstall the messaging apps. \ours performs well without extensive reasoning.}
    \label{fig:enter-label}
\end{figure}

\begin{table*}[h!]
  \begin{minipage}{0.99\textwidth}
\centering
\captionof{table}{
Counting example from SuperCLEVR~\cite{li2023super}. Although the question is out-of-distribution, the performance of \ours and vanilla GRPO differs significantly in their outputs. The output from vanilla GRPO is excessively lengthy (over 500 tokens), focusing more on reasoning rather than providing a direct answer. In contrast, \ours delivers a concise response, effectively bypassing the lengthy reasoning process.}

\scalebox{0.88}{
\begin{tabular}{l p{5.4cm} p{6.0cm} }
\toprule
 \multicolumn{3}{l}{\bf Counting examples from SuperCLEVR~\cite{li2023super}}  \\
\midrule
&  \includegraphics[height=3.5cm]{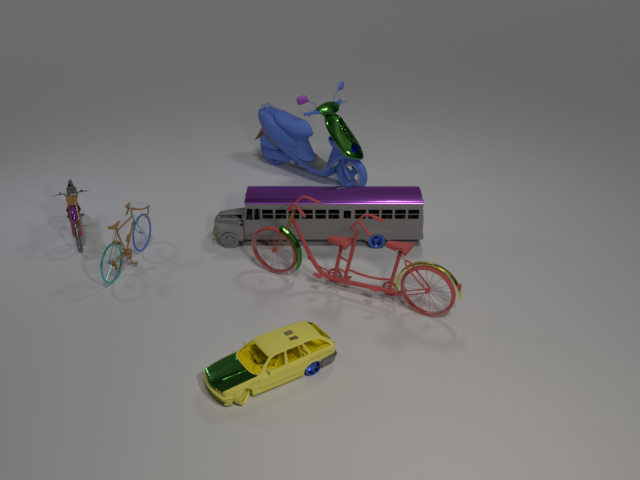} 
&  \includegraphics[height=3.5cm]{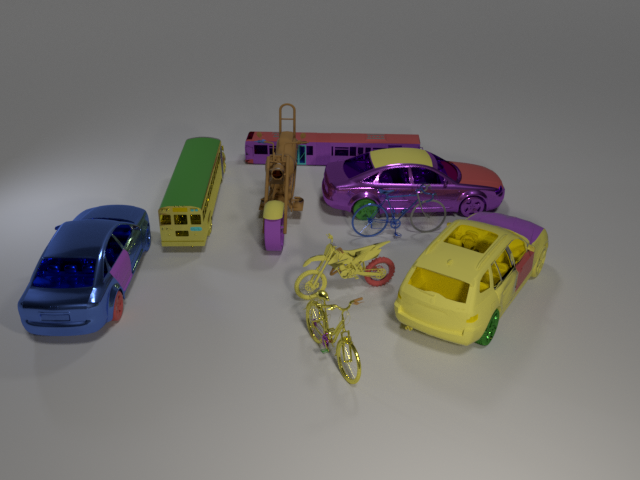} \\
\midrule
Question & How many objects are either small green wagons or buses?
 & How many other objects are there of the same color as the matte car?
\\ \midrule
GRPO's outputs & 

"<think>Okay, let's tackle this question step by step. The user is asking how many objects are either \"small green wagons\" or \"buses\". I need to check each object in the image provided.First, let's parse the image. There are six objects listed:1. A large blue rubber scooter with specific coordinates and rotation.2. A small yellow rubber car (wagon) with details.3. A small brown rubber bicycle.4. A large red rubber tandem bicycle.5. A small gray metal dirt bike.6. A large gray metal school bus.The question is about two categories: small green wagons and buses. Let's break down each part.Starting with \"small green wagons\": Looking at the list, there's a \"small yellow rubber car\" which is a wagon. The color here is yellow, not green. So that doesn't fit. Are there any other small green wagons? The other objects are all different colors and types. The only green object mentioned is the first one, but it's a scooter, not a wagon. So no small green wagons exist in the scene.Next, \"buses\": The sixth object is a \"large gray metal school bus\". Since the question says \"",

& 
<think>Okay, let's see. The question is asking how many other objects are there of the same color as the \"matte car\". First, I need to figure out which object is the matte car.Looking at the image, each object has a color mentioned. Let me check each entry:1. A large yellow rubber SUV...2. A small blue metal bicycle...3. A large purple metal sedan...4. A large brown metal cylinder...5. A large yellow metal school bus...6. A small yellow metal bicycle...7. A large purple metal sedan...8. A small yellow rubber dirt bike...9. A large yellow metal school bus...10. A large purple metal articulated train...Wait, the matte car isn't listed here. All the colors mentioned are either yellow, blue, purple, brown, or gray. The matte material isn't specified for any of them. Hmm, maybe \"matte\" refers to the material being non-shiny, but in the given data, all materials are either metal or rubber. So perhaps the matte car is one of the rubber ones. But the problem says \"matte\", so maybe it's a mistake, or maybe the user intended \"rubber\" instead of \"matte",

\\ \midrule
TON's outputs & <think>\textbackslash n\textbackslash n</think>

<answer>1</answer> & 
<think>\textbackslash n\textbackslash n</think>

<answer>3</answer>
\\ \bottomrule
\end{tabular}
}

\label{tab:clevr example}  
  \end{minipage}
\end{table*}
ewpage
\begin{figure}
    \centering
    \includegraphics[width=\linewidth]{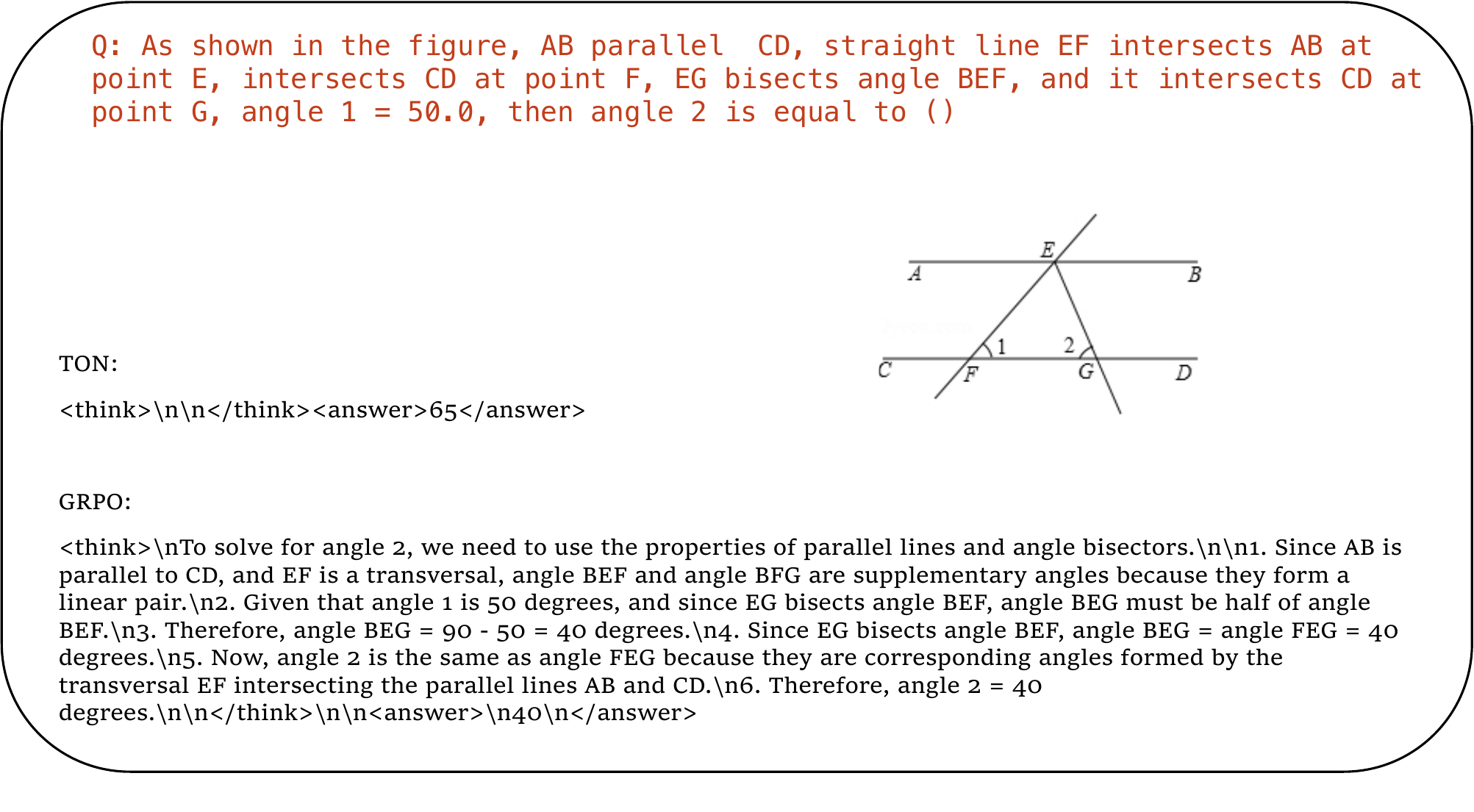}
    \caption{Comparison between \ours and vanilla GRPO in GeoQA}
    \label{fig:enter-label}
\end{figure}

\begin{figure}
    \centering
    \includegraphics[width=\linewidth]{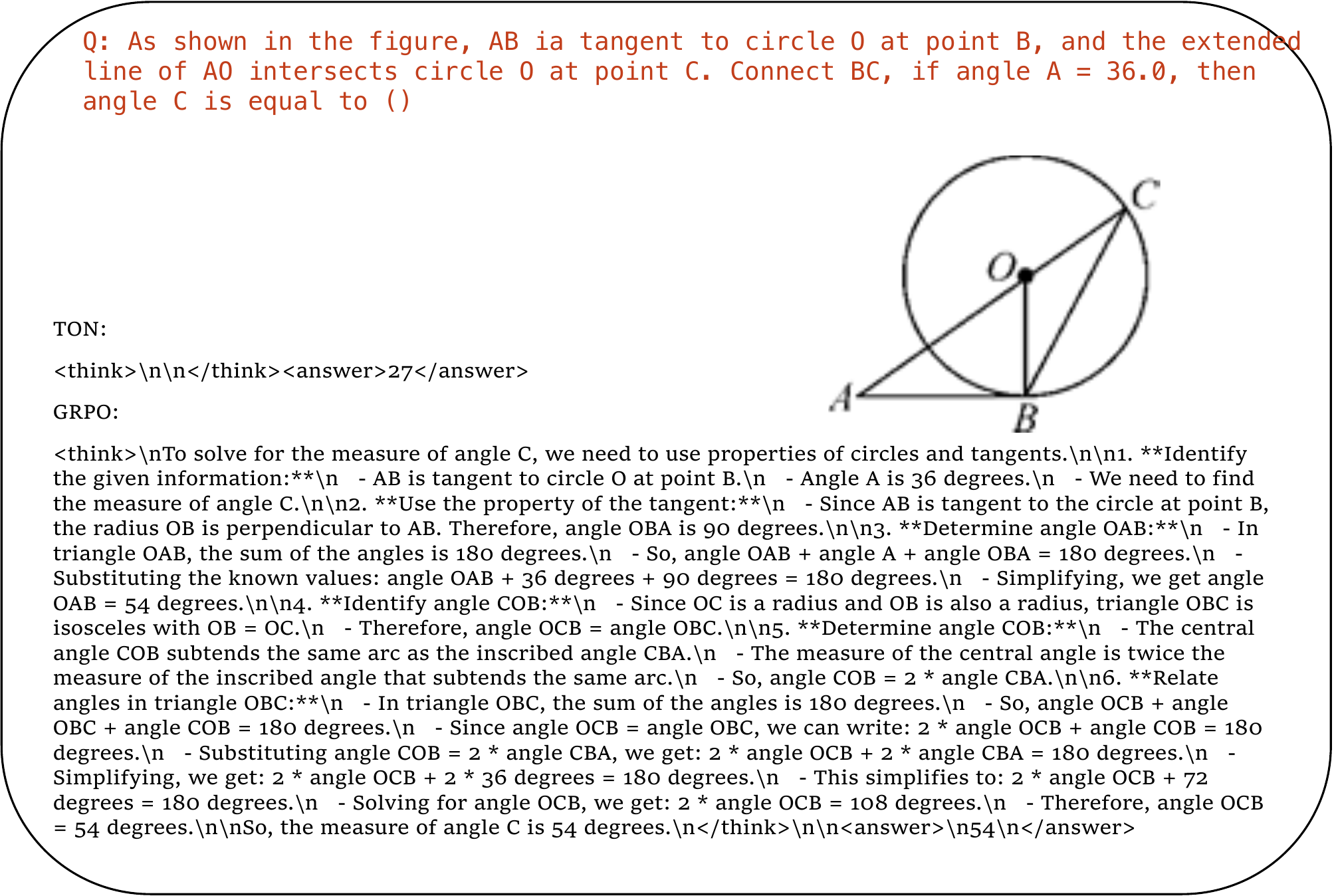}
    \caption{Comparison between \ours and vanilla GRPO in GeoQA}
    \label{fig:enter-label}
\end{figure}

\begin{figure}
    \centering
    \includegraphics[width=\linewidth]{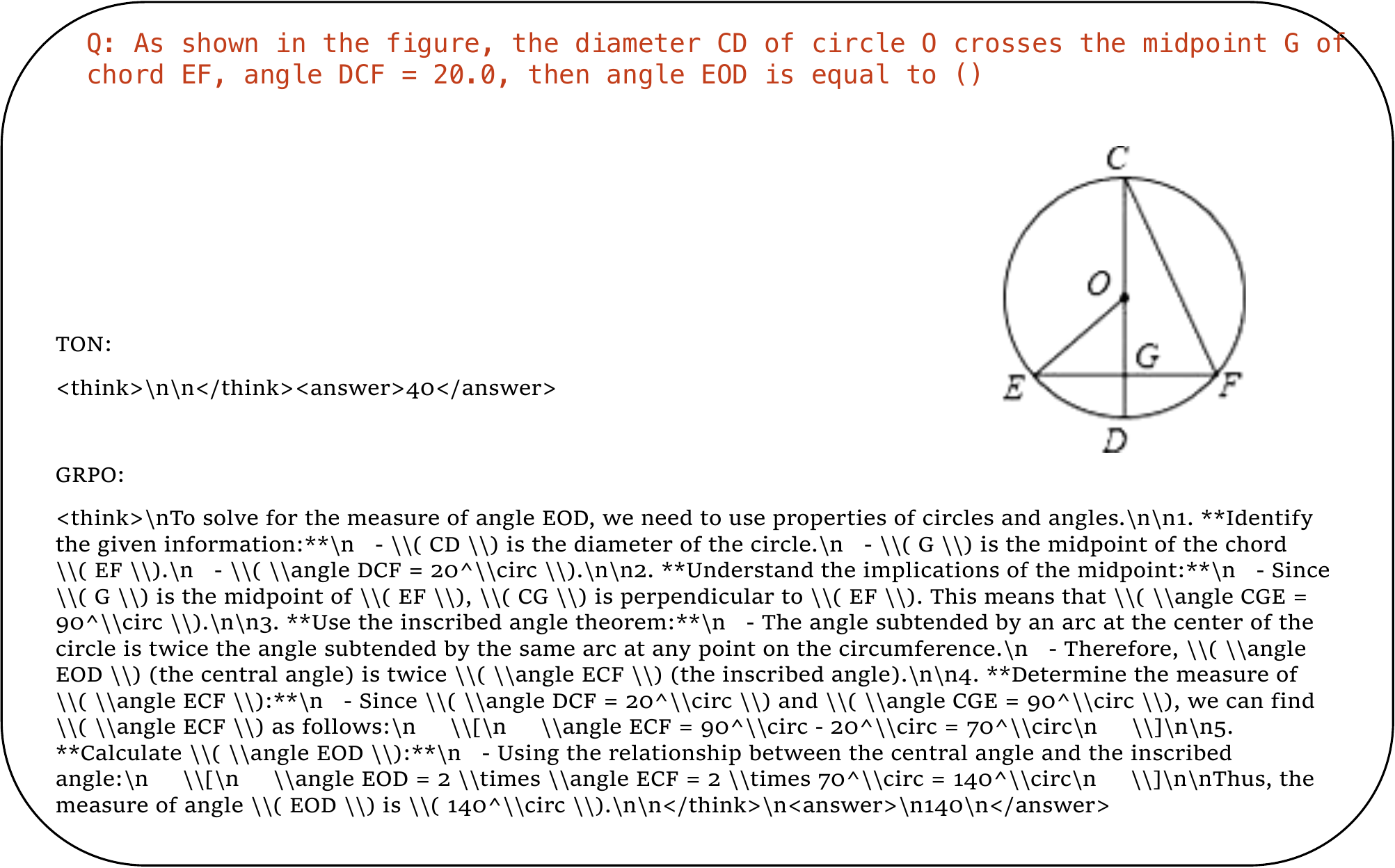}
    \caption{Comparison between \ours and vanilla GRPO in GeoQA}
    \label{fig:enter-label}
\end{figure}

\begin{figure}
    \centering
    \includegraphics[width=\linewidth]{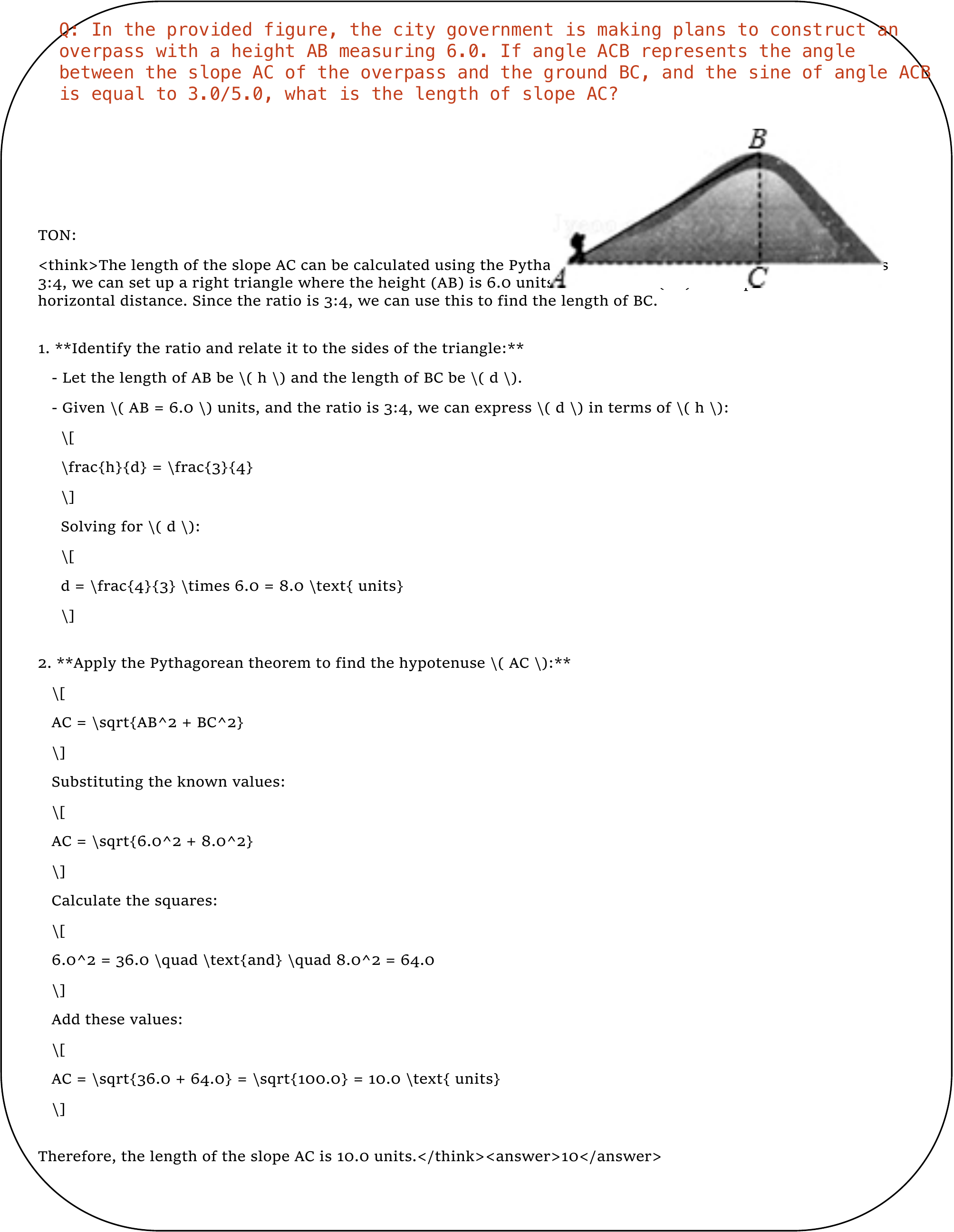}
    \caption{Comparison between \ours and vanilla GRPO in GeoQA}
    \label{fig:enter-label}
\end{figure}

\begin{figure}
    \centering
    \includegraphics[width=\linewidth]{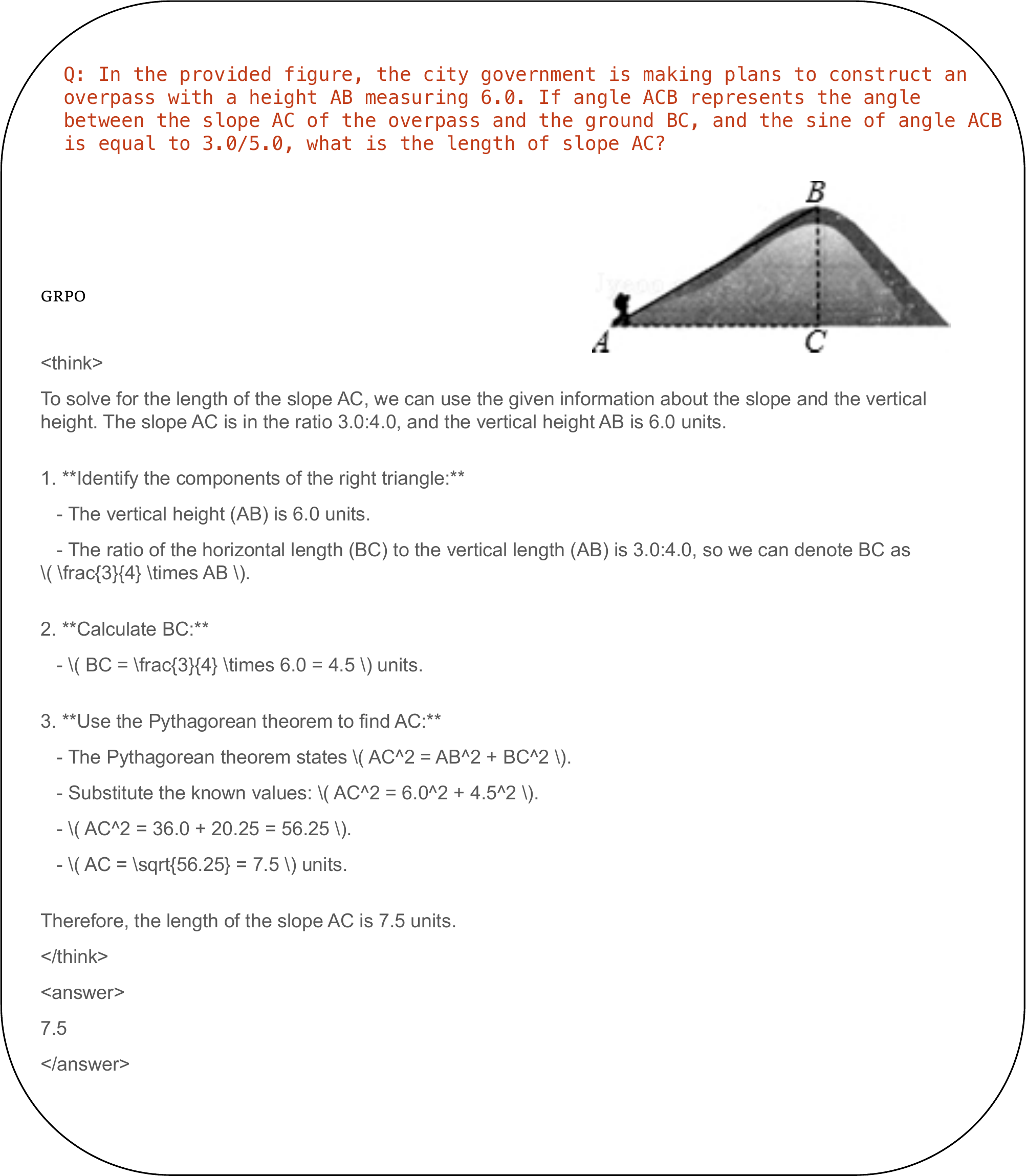}
    \caption{Comparison between \ours and vanilla GRPO in GeoQA}
    \label{fig:enter-label}
\end{figure}